\documentclass[iicol]{sn-jnl}
\jyear{2026}%

\theoremstyle{thmstyleone}
\theoremstyle{thmstyletwo}

\theoremstyle{thmstylethree}

\usepackage{graphicx}
\usepackage{amsmath}
\usepackage{amssymb}
\usepackage{booktabs}
\usepackage{comment}
\usepackage{multirow}
\usepackage{mathtools}
\usepackage{epsfig}
\usepackage[misc]{ifsym}

\usepackage{makecell}
\usepackage{pifont}
\usepackage{bbding}
\usepackage{wrapfig}
\usepackage{verbatim}
\usepackage{relsize}
\usepackage{colortbl}
\usepackage{bm}
\usepackage{bbm}
\usepackage{breakurl}
\usepackage{color}
\usepackage{diagbox}
\usepackage{amsmath}
\usepackage{multirow}
\usepackage[normalem]{ulem}
\useunder{\uline}{\ul}{}
\usepackage{tabularx}
\usepackage{multirow}
\usepackage{algorithm}%
\usepackage{algorithmicx}%
\usepackage{algpseudocode}%
\usepackage{threeparttable}
\usepackage{booktabs}
\usepackage[capitalize]{cleveref}
\usepackage{fancyvrb,cprotect}

\usepackage[numbers]{natbib} 

\DeclareUnicodeCharacter{2212}{-}

\begin{document}
	
	\title[Article Title]{Intervention-Based Self-Supervised Learning: A Causal Probe Paradigm for Remote Photoplethysmography}
	
	\author[1,2]{\fnm{Zhiyi} \sur{Niu}}\email{ZhiyiNiu@outlook.com}
	\author[2]{\fnm{Xiaoguang} \sur{Tu}}\email{xguangtu@outlook.com}
	\author[3]{\fnm{Bo} \sur{Zhao}}\email{225040248@link.cuhk.edu.cn}
	\author[3]{\fnm{Junzhe} \sur{Cao}}\email{caojunzhe\_hitsz@stu.hit.edu.cn}
	\author*[1]{\fnm{Dan} \sur{Guo}}\email{guodan@hfut.edu.cn}
	\author*[3]{\fnm{Zitong} \sur{Yu}}\email{yuzitong@gbu.edu.cn}
	
	\affil[1]{\orgname{School of Computer Science and Information Engineering, Hefei University of Technology, Hefei 230601, PR China}}
	
	\affil[2]{\orgname{College of Aviation Electronic and Electrical Engineering, Civil Aviation Flight University of China, Guanghan 618307, PR China}}
	
	\affil[3]{\orgname{School of Computing and Information Technology, Great Bay University, Dongguan 523000, PR China}}

	\abstract{Remote Photoplethysmography (rPPG) enables convenient non-contact physiological measurement. Existing Self-Supervised Learning (SSL) methods commonly fall into a correlation trap: they tend to learn the most dominant periodic signals in the data, such as high-energy motion or illumination noise, rather than the faint, true rPPG signal, leading to poor model generalization. To address this, we propose a new SSL paradigm, Physiological Causal Probing (PCP), which treats the latent rPPG signal as the underlying physical source and the resulting pixel chrominance variations as its visual manifestation. Its core idea is to shift from passive correlation learning to active, precise intervention: it intervenes on the video based on a proposed rPPG hypothesis, and verifies whether the post-intervention changes match physical expectations. We propose the Interv-rPPG framework to implement PCP: an rPPG extractor named PhysMambaFormer hypothesizes the rPPG signal, while a Controllable Physiological Signal Editor conducts precise chrominance-domain interventions on videos based on this hypothesis. Interv-rPPG validates the physical realism of the hypothesis through `Falsifiability via Nulling' and `Axiomatic Equivariance'. Our editor achieves precise editing of the rPPG signal by intervening in the low-frequency chrominance components of the video. Our method improves both in-domain and cross-domain performance on challenging datasets such as VIPL-HR and MMPD. Furthermore, it surpasses the supervised baseline in complex cross-dataset settings, while remaining competitive on clean datasets where the intervention mechanism may introduce slight residual chrominance noise. Extensive experiments, including diagnostic analysis of nuisance sensitivity, demonstrate that the PCP paradigm effectively resists motion and illumination artifacts.
	}

	
	\maketitle
	
	\vspace{-1.5em}
	\section{Introduction}
	\vspace{-0.3em}
	
	\thispagestyle{empty}

	Remote Photoplethysmography (rPPG) is revolutionizing fields such as health monitoring, telemedicine, and affective computing by enabling non-contact measurement of vital physiological indicators like heart rate and heart rate variability through the analysis of subtle color changes in facial videos captured by ordinary cameras \cite{de2013robust, ru2026unveiling}. The potential for popularizing this technology is immense, as it can transform ubiquitous consumer-grade cameras into powerful physiological sensors, paving the way for large-scale, low-cost public health screening \cite{chen2024deep, zhao2025phase}.

	However, the rPPG field has long faced a major bottleneck: a critical dependence on large-scale, diverse, and precisely synchronized labeled data. Collecting such datasets is not only expensive and labor-intensive but also fundamentally incomplete, as it can never encompass the infinite variations of real-world conditions, including different lighting, skin tones, motion patterns, and device characteristics \cite{pirzada2023remote, pirzada2024remote}. Consequently, robust domain generalization remains a key goal for researchers. This fundamental challenge has significantly spurred research into self-supervised learning (SSL) methods \cite{park2022self, wang2026micro, wang2022self, canellas2025self, speth2023non, speth2024sinc+}, which aim to autonomously learn intrinsic representations of the rPPG signal from vast amounts of unlabeled video data.

	Despite the promise of SSL in addressing data scarcity, existing paradigms like contrastive learning \cite{gideon2021way,sun2022contrast}, generative models \cite{liu2024rppg}, and pseudo-labeling techniques \cite{li2023contactless, du2021weakly, xie2024sfda} remain dominated by observational correlation. In rPPG, this limitation is particularly severe because the target signal is extremely weak: the pulse appears mainly as subtle skin chrominance fluctuations, whereas head motion, ambient light flicker, and compression artifacts often generate periodic patterns with much larger energy and stronger temporal regularity. Under such weak-signal conditions, observation alone cannot determine whether a periodic component is physiologically valid or merely dominant in magnitude. As a result, existing SSL models often learn nuisance-dependent representations that transfer poorly across datasets.

		\begin{figure}[t]
	\centering
	\includegraphics[scale=0.3]{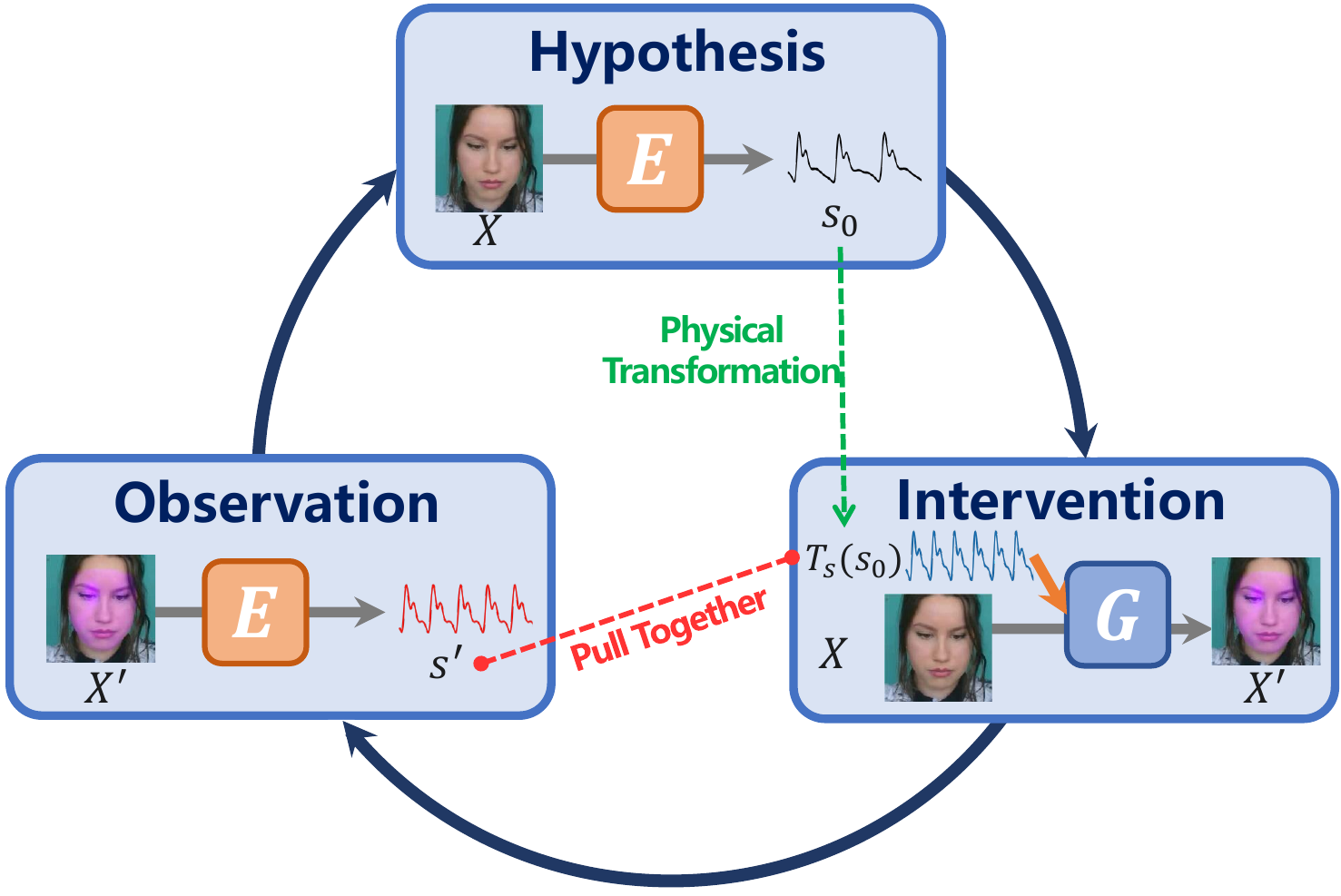}
	\vspace{-0.5em}

		\caption{\small{
				Visualization of the Physiological Causal Probing Paradigm. The extractor ($E$) proposes an rPPG signal hypothesis $s_0=E(X)$ from the original video $X$, then applies a physical transformation on $s_0$ to obtain $T_s(s_0)$, and the signal editor ($G$) performs interventions on $X$ based on $T_s(s_0)$ to generate the intervened video $X^{'}$. In the observation phase, the extractor outputs $s^{'}=E(X^{'})$, and the verification is performed by explicitly comparing $s^{'}$ with the expected transformed signal $T_s(s_0)$.}
		}
	\label{fig:PCP}
\end{figure}

		To break this impasse, we move from passive correlation learning to intervention-based identification. Physiological Causal Probing (PCP) is built on the physical formation pathway of rPPG: pulse-induced blood-volume changes give rise to subtle chrominance fluctuations on facial skin. As illustrated in Fig.~\ref{fig:PCP}, the extractor $E$ first proposes a hypothetical signal $s_0$ from the original video $X$, then applies a physical transformation on $s_0$ to obtain $T_s(s_0)$. The editor $G$ then perturbs $X$ along this chrominance pathway according to $T_s(s_0)$ to produce the intervened video $X^{'}$, and the extractor finally checks whether the post-intervention signal follows the expected transformation. Formally, the observation step compares the extracted post-intervention signal $s^{'}=E(X^{'})$ against the expected transformed hypothesis $T_s(s_0)$; the verification target is therefore $s^{'} \approx T_s(s_0)$. A signal that remains consistent under such controlled intervention is more likely to be the true physiological signal, whereas motion- or illumination-driven periodicities may correlate with the video but are not governed by the same editable chrominance mechanism. This hypothesis–intervention–verification loop provides the physical grounding that conventional observation-only SSL lacks.

	We introduce the Interv-rPPG framework as a concrete implementation of the PCP paradigm. The framework consists of an extractor named PhysMambaFormer and a Controllable Physiological Signal Editor. PhysMambaFormer is capable of rapid and accurate rPPG signal extraction, while the editor specializes in precise modification of rPPG signals within videos. Specifically, PhysMambaFormer is responsible for proposing rPPG signal hypothesis and the editor performs signal interventions on the video. The framework is built upon two theoretical pillars:

	\noindent\textbf{Axiomatic Equivariance:}\quad  
	This principle mandates that the learned representation must adhere to the equivariance axioms of signal physics. When the physical properties of a signal—such as its amplitude, phase, or frequency—are altered, its visual manifestation in the video should change correspondingly. Any video artifact that does not conform to these physical properties will be filtered out by this constraint.

	\noindent\textbf{Falsifiability via Nulling:}\quad  
	The model actively attempts to falsify its own predictions. Specifically, the extractor hypothesizes a signal to be the true rPPG signal, the editor injects an `anti-signal' of equal amplitude and opposite phase into the video. If the hypothesized signal is genuine, it should be physically canceled, causing the signal energy in the intervened video to approach zero. If it is an artifact like motion, this color-based intervention will fail to nullify it, thus falsifying the initial hypothesis.

	This closed-loop logic of `hypothesis-intervention-observation' compels the extractor to learn a representation that is causally linked to the physical phenomenon of rPPG, rather than merely statistically correlated. Our main contributions can be summarized as follows:
	
	\begin{itemize}
		\setlength\itemsep{-0.1em}
		\vspace{-0.5em}
		\item We introduce the Physiological Causal Probing (PCP) paradigm to the self-supervised rPPG field, shifting the learning process from traditional correlation modeling to a new paradigm based on hypothesis–intervention–verification loop.

		\item  We propose Interv-rPPG, a framework grounded in Axiomatic Equivariance and Falsifiability via Nulling, to implement the PCP paradigm, providing a powerful physical anchor and self-correction mechanism for self-supervised learning.

		\item We design a novel Controllable Physiological Signal Editor capable of precise and controllable editing of rPPG signals in video. This is not only the core component of our framework, but also a key advancement beyond existing rPPG signal editing methods (e.g., PulseEdit) that merely apply global perturbations.

		\item We conduct rigorous intra-domain and cross-domain testings on multiple datasets, demonstrating that Interv-rPPG not only achieves good performance in intra-domain testings but also shows superior results in the challenging cross-dataset generalization testings, addressing the critical issue of robustness in the field.

	\end{itemize}
	
	In the rest of the paper, Section.~\ref{sec:relatedwork} provides the related work about rPPG measurement and Self-Supervised Learning methods. Section.~\ref{sec:method} introduces the PCP paradigm and Interv-rPPG framework. Section.~\ref{sec:experiemnts} introduces the results of rPPG measurement, video physiological signal
	editing evaluation, and rigorous ablation studies. Finally, conclusions  and future work are given in Section.~\ref{sec:conc}.

	\section{Related Work}
	\label{sec:relatedwork}
	
	\subsection{Remote Physiological Measurement}  
	Remote Photoplethysmography enables the remote monitoring of cardiac activity by analyzing minute, pulse-induced color changes on the skin surface in facial videos. 
	
	\noindent\textbf{rPPG signal extraction methods.}\quad 
	In-depth research has led to a variety of rPPG signal extraction methods. Early approaches were based on optical reflection principles and signal separation theory, using handcrafted features. The GREEN \cite{verkruysse2008remote} method proposed directly using the green channel signal from a facial ROI, leveraging the strong absorption of green light by hemoglobin. ICA \cite{poh2010advancements} treated the three RGB channel signals as a mixed source, using blind source separation to extract statistically independent components and selecting the one corresponding to the pulse frequency. CHROM \cite{de2013robust} enhanced signal robustness by applying white balance and using a linear combination of chrominance channels to cancel illumination variations. POS \cite{wang2016algorithmic} defined a plane orthogonal to the skin tone in a normalized RGB space, projecting the signal to separate the pulse from motion and illumination noise.
	
	With the rise of deep learning, DeepPhys \cite{chen2018deepphys} was the first to apply CNNs for rPPG extraction, while PhysNet \cite{yu2019remote1} used a 3D-CNN autoencoder to reconstruct the rPPG waveform. TS-CAN \cite{liu2020multi} integrated multi-scale temporal convolutions with attention mechanisms to improve adaptation to dynamic scenes. PhysFormer \cite{yu2022physformer}, based on a Temporal Difference Transformer, enhanced robustness by capturing long-range spatio-temporal dependencies. Dual-GAN \cite{lu2021dual} employed a dual generative adversarial network structure to suppress interference, and PhysMamba \cite{luo2024physmamba} utilized a Mamba \cite{gu2024mamba} (State Space Model) architecture with a SlowFast design to capture both local dynamics and long-term context. Qian et al. \cite{qian2025physdiff} proposed PhysDiff, a diffusion-based model for rPPG estimation that innovatively incorporates physiology-inspired dynamicity disentanglement, along with a spatio-temporal hybrid denoiser and periodicity-based multi-hypothesis selection, greatly enhancing estimation robustness and cross-domain generalization in complex interference scenarios.

	\noindent\textbf{Editing Physiological Signals in Videos.}\quad 
	Due to the difficulty and high cost of rPPG data collection, editing physiological signals in existing videos has emerged as an efficient and low-cost method for data augmentation. Early work by Chen et al. \cite{chen2017eliminating} used Motion Component Magnification (MCM) to locate and suppress motion components related to the physiological signal. Later, to amplify signals, Li et al. \cite{li2023hiding} superimposed sinusoidal perturbations onto facial ROI pixels. Moving beyond simple erasure, Chen et al. \cite{chen2022pulseedit} proposed PulseEdit, which calculates the difference between a target rPPG signal and the actual pixel fluctuations and adds this difference as a perturbation to achieve directional editing.
	
	However, these pixel-level editing methods often create a conflict between achieving precise signal editing and preserving the high-fidelity of the original video content. To address this, generative models were introduced. Tu et al. proposed Phys-EdiGAN \cite{tu2025phys}, which uses a conditional GAN to generate a video containing a given rPPG signal as a condition. Zhou et al. introduced PhysioLatent \cite{zhou2025editing} for editing signals in the video's latent space. A key drawback of these generative approaches is that they must focus not only on generating the specified rPPG features but also on reconstructing all other video information, leading to unnecessary training overhead and difficult-to-control results.
	
	In conclusion, pixel-level perturbation methods struggle to balance signal accuracy with video fidelity, while generative methods are computationally expensive and cannot focus solely on modifying the rPPG feature domain. Therefore, the problem of how to achieve both precise rPPG signal editing and a lightweight, efficient model remains a critical challenge to be solved.

	\vspace{0.5em}
	\subsection{Self-Supervised Learning.}
	Self-supervised learning is a machine learning method that does not require manual labeling of data and trains models by mining supervised signals from the data itself. It greatly alleviates the problem of data scarcity during model training.
	
	\noindent\textbf{Contrastive Learning.}\quad  
	Contrastive learning \cite{he2020momentum, zhang2026multimodal, misra2020self, chen2025contrastive} is one of the most popular SSL paradigms, which core idea is to learn an embedding space where similar samples (positive pairs) are pulled closer together and dissimilar samples (negative pairs) are pushed farther apart. Gideon et al. \cite{gideon2021way} constructed positive and negative pairs through temporal resampling, which essentially trains the model to separate the strongest periodic signals in the video. This strategy may be effective in a laboratory setting with clean data and minimal interference, but in real-world scenarios, the energy and consistency of periodic noise often far exceed the weak rPPG signal. In contrast, Contrast-Phys+ \cite{sun2024contrast} no longer blindly assumes that the dominant signal is the target signal. Instead, it cleverly incorporates physical priors such as spatial similarity, temporal similarity, and cross-video variability of rPPG signals to constrain the contrastive learning process. The model is constrained to find a signal that remains consistent across different facial regions and over short periods of time. However, the essence of this learning framework remains passive and correlation-based. It can still be deceived by noise that happens to mimic these physical priors. For example, when a person shakes their head rhythmically, or when there is periodic flickering in environmental lighting, the signals also exhibit spatial and temporal consistency. These methods may still be misled to learn stronger noise signals.
	
	\noindent\textbf{Pseudo-Labeling Assisted Methods.}\quad  
	Recognizing that contrastive methods fail under interference from irrelevant periodic noise, Li et al. \cite{li2023contactless} proposed using non-learning traditional methods to generate pseudo-labels to guide model training, along with a label-refinement technique to mitigate the impact of noisy labels. This approach suffers from a ceiling effect, as its performance is fundamentally capped by the efficacy of the chosen traditional method \cite{wang2015novel}, severely limiting its potential.
	
	\noindent\textbf{Generative and Cycle-Consistency Methods.}\quad  
	Another SSL methods, such as Re-rPPGNet \cite{hsieh2022augmentation}, utilizes generative models and cycle-consistency losses. These methods aim to learn to remove or embed rPPG signals from video, relying on cycle-consistency \cite{zhu2017unpaired} to maintain visual realism. However, the theoretical underpinning of this approach rests merely on the reversibility of the operation, not its physical correctness. Given the extremely low energy of the rPPG signal in video, a model can learn to cheat the loss by adding and removing a visually imperceptible yet non-physiological artifact. As long as the operation is reversible, the cycle-consistency loss is perfectly satisfied, but the model learns nothing about the true rPPG signal. This creates a logically closed loop that lacks any connection to the physical essence of rPPG.

	\noindent\textbf{Reconstruction Methods.}\quad  
	To enhance the understanding of global semantic information, He et al. \cite{he2022masked} first introduced the masked-and-reconstruct paradigm (MAE) to computer vision, learning deep features from reconstructing missing information. Liu et al. \cite{liu2024rppg} adapted MAE to the rPPG domain, seeking to enhance the model's perception of the rPPG signal by reconstructing missing parts of a video's Spatio-Temporal Map (STMap) \cite{niu2019rhythmnet}. However, the supervision signal in MAE-based methods comes from the pixels themselves. To minimize reconstruction error, the model is strongly incentivized to learn information-rich surface textures and illumination patterns rather than the extremely low-energy rPPG signal.

	\vspace{0.5em}
	In summary, all SSL paradigms currently applied in the rPPG field can essentially be attributed to correlation learning. Their supervisory signals all originate from the statistical properties of the data distribution itself, such as self-similarity, reversibility, or reconstructability. This fundamental reliance on statistical correlation makes it difficult to stably distinguish low-energy physiological signals from high-energy noise in uncontrolled environments. To overcome this bottleneck, we propose a fundamental paradigm shift: from passively observing statistical correlations in data to actively exploring their inherent causal relationships through physical intervention.

	\begin{figure}[t]
		\centering
		\includegraphics[scale=0.32]{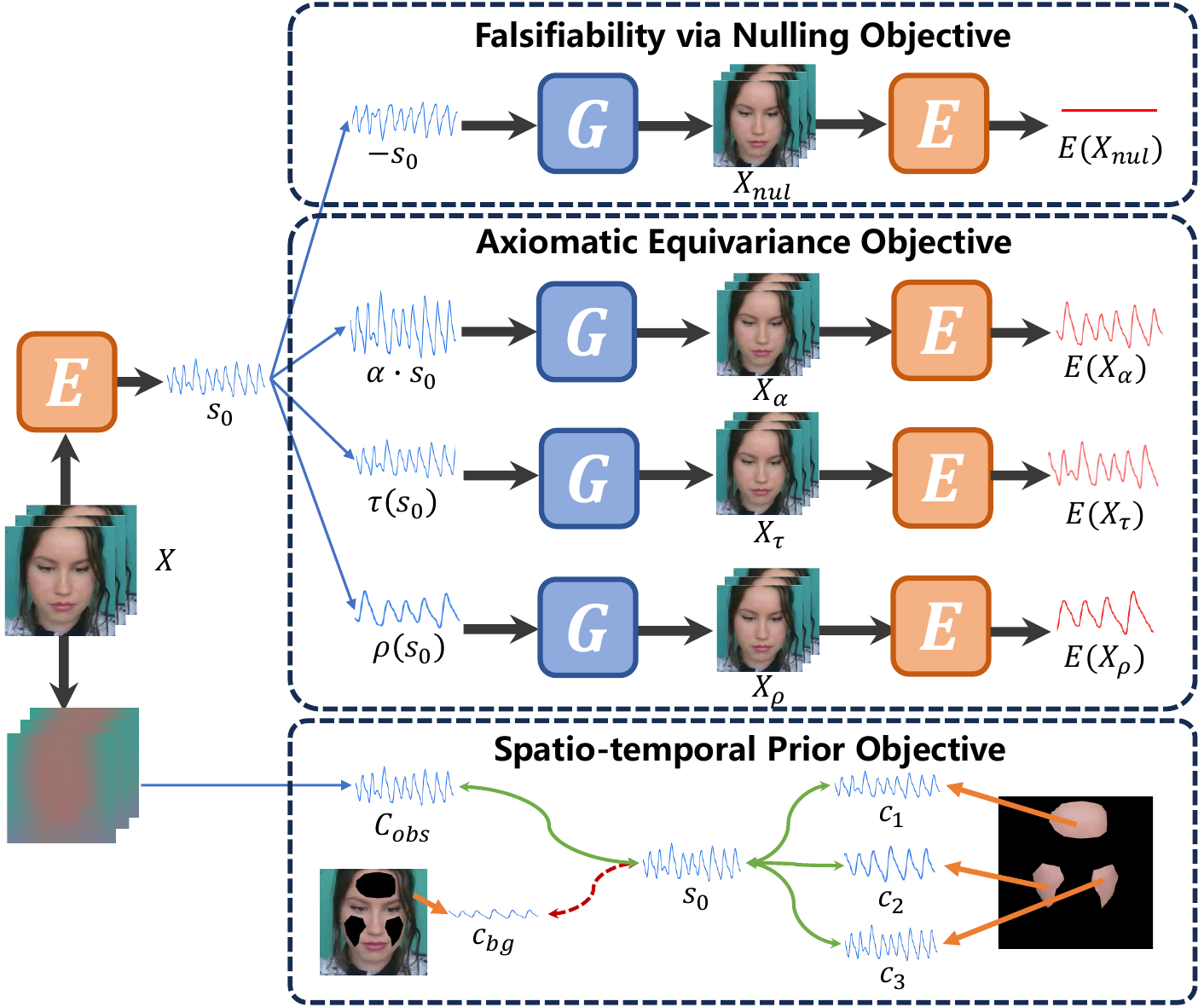}
		\vspace{-0.5em}
		\caption{\small{
				The three major objectives in the PCP paradigm: (1) Falsifiability via Nulling Objective, (2) Axiomatic Equivariance Objective, and (3) Spatio-temporal Prior Objective.}
		}
		\label{fig:Object}
	\end{figure}

	\section{Methodology}
	\label{sec:method}
	This section introduces a novel self-supervised paradigm, termed Physiological Causal Probing (PCP). We then detail its implementation through a new framework, Interv-rPPG, which is composed of an rPPG extractor (PhysMambaFormer) and a Controllable Physiological Signal Editor. 
	
	\subsection{Physiological Causal Probing Paradigm}
	\label{sec:PCP}
	
	As previously discussed, existing self-supervised methods mainly learn from statistical regularities such as consistency \cite{gideon2021way, speth2023non, sun2024contrast}, reversibility \cite{hsieh2022augmentation}, and reconstructability \cite{liu2024rppg}. PCP instead formulates rPPG learning as a physically testable hypothesis–intervention–verification process.

	Within the PCP paradigm, we construct a physiological signal editor ($G$) and a rPPG extractor ($E$) as its core components. The editor can precisely manipulate the rPPG signal within a facial video, altering physical properties such as amplitude, frequency, and phase. The extractor is responsible for predicting the rPPG signal from the video. PCP paradigm first uses the extractor to formulate a hypothesis about the rPPG signal. It then employs the editor to intervene on the video based on this hypothesis. If the rPPG signal extracted from the intervened video exhibits a corresponding change, it provides evidence that the extractor has learned a true representation of the rPPG signal. As shown in Fig.~\ref{fig:Object}, the PCP paradigm consists of three major objectives: (1) Falsifiability via Nulling Objective, (2) Axiomatic Equivariance Objective and, (3) Spatio-temporal Prior Objective.

	More concretely, PCP accepts a representation only when it passes an intervention test: if a predicted signal is truly physiological, then chrominance nulling should suppress it and a transformed hypothesis should reappear as the corresponding transformed signal after intervention. This requirement is stricter than ordinary consistency because it demands both editability in the video domain and predictability in the signal domain. Therefore, PCP does not merely ask whether a periodic component is present. It asks whether the component behaves like the physical consequence of blood-volume-induced chrominance modulation.

	\begin{figure*}[t]
		\centering
		\includegraphics[scale=0.37]{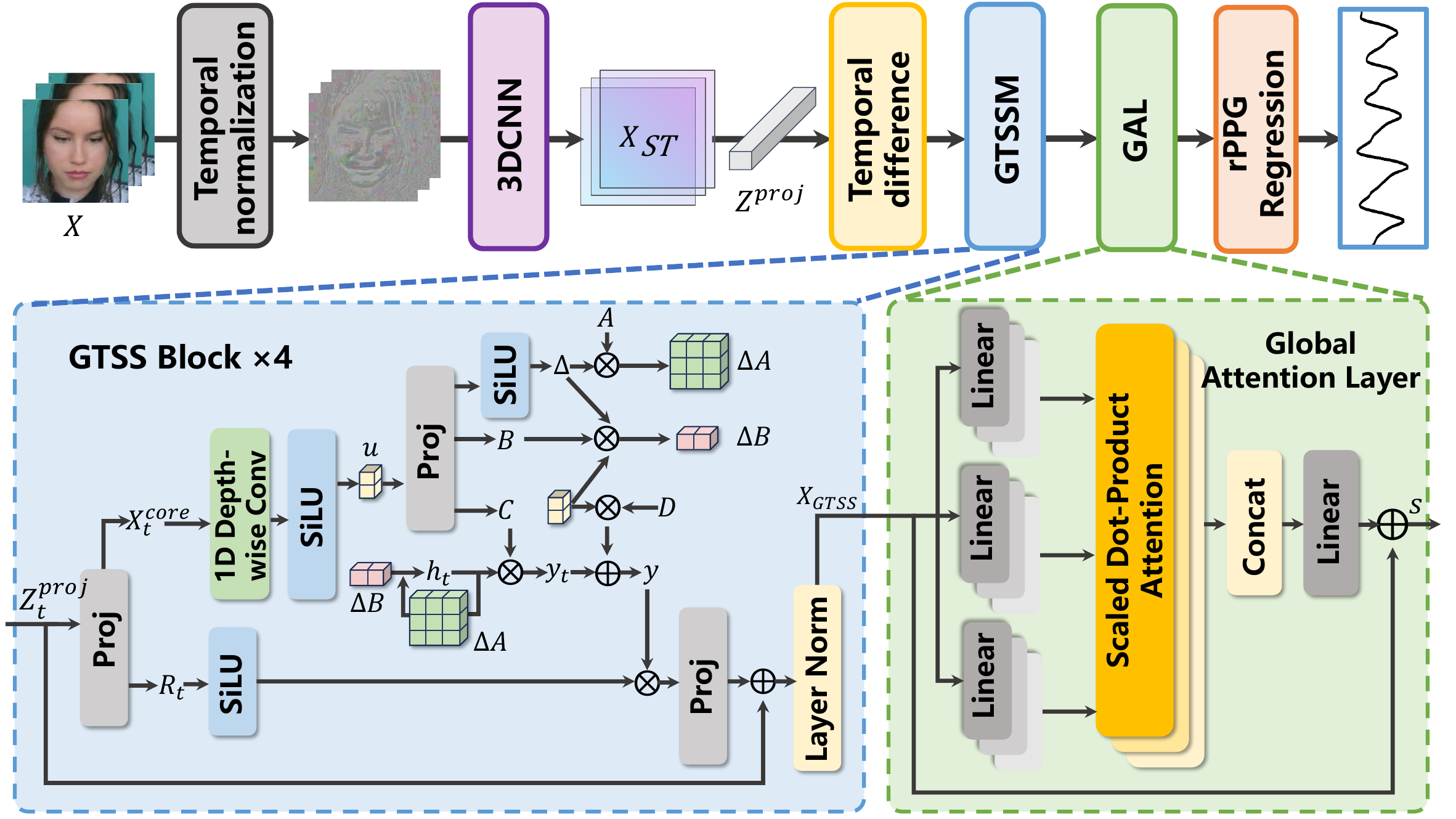}
		\vspace{-0.3em}
		\caption{\small{
				Framework of the PhysMambaFormer. The input video ($X$) is first temporally normalized by per-pixel detrending and variance normalization along the time axis, and a lightweight 3D-CNN extracts local spatio-temporal features ($X_{ST}$). The flattened per-frame features are projected to latent tokens($Z_t^{proj}$), and a temporal-difference module is added to emphasize pulse-synchronous temporal variations relative to slow drift. The temporal backbone consists of 4 Gated Temporal State Space (GTSS) blocks. In each GTSS block, the token $Z_t^{proj}$ is split into a main branch ($X_t^{core}$) and a gating branch ($R_t$); the state-space parameters are dynamically generated from $X_t^{core}$. The output is further refined by a Global Attention Layer (GAL), and a linear regression head outputs one rPPG sample per frame.
			}
		}
		\label{fig:PhysMambaFormer}
		\vspace{-0.3em}
	\end{figure*}

	\vspace{0.3em}
	\noindent\textbf{Falsifiability via Nulling Objective}\quad  
	
	A hypothesis is scientific only if it is falsifiable. In PCP, the extractor's output for a video $X$, denoted as $s_0 = E(X)$, is treated as a hypothesis that $s_0$ is the true rPPG signal contained within video $X$. We design a falsifiability experiment: using the editor $G$, we inject an `anti-signal' $-s_0$ (with equal amplitude and opposite phase) into the video domain to obtain a null-pulse video $X_{nul} = G(X, -s_0)$. This intervention attempts to physically nullify the signal in the chrominance domain.
	
	This leads to a falsifiability criterion: If extractor's hypothesis is correct (i.e., $s_0$ is the true rPPG component that can be linearly represented by chrominance), the editor $G$ should effectively cancel it. The signal extracted from the nullified video, $E(X_{nul})$, should then have energy approaching zero. Conversely, if extractor's hypothesis is incorrect (e.g., $s_0$ is an artifact of head motion or illumination changes), the editor $G$ which operates only on chrominance level, will fail to cancel these non-chrominance-based artifacts. Significant residual energy will remain in $X_{nul}$. In this case, the incorrect hypothesis is successfully falsified.

	\vspace{0.3em}
	\noindent\textbf{Axiomatic Equivariance Objective.}\quad  
	Equivariance is a fundamental concept in physics and machine learning, requiring that when an input undergoes a transformation, the system's output should undergo a corresponding, predictable transformation. We introduce this principle into rPPG self-supervised learning as a physical axiom that the model must obey.

	Formally, let $T_v(\cdot)$ be an interventional transformation applied to the rPPG component in the video domain, and let $T_s(\cdot)$ be the corresponding transformation in the signal domain. We aim to ensure the extractor $E$ satisfies the equivariance constraint: 
	
	\vspace{-0.9em}
	\begin{equation}
		E(T_v(x)) \approx T_s(s_0).
	\end{equation}
	
	This means the signal extracted after intervening on the video should be approximately equal to the signal extracted before the intervention and then transformed. We define three key interventions: (1) amplitude, (2) phase, and (3) frequency.

	\textbf{(1) Amplitude Intervention}
	
	This intervention uses the editor $G$ to scale the rPPG signal's amplitude in the video. The corresponding signal-domain transformation is a linear scaling by a factor $\alpha$. The formula is:
	
	\vspace{-0.9em}
	\begin{equation}
		\begin{aligned}
			&X_\alpha = G(X, \alpha \cdot s_0),\\
			&E(X_\alpha) \approx (1 + \alpha) \cdot s_0.
		\end{aligned}
	\end{equation}
	
	This forces the model to learn a representation with a linear response to the rPPG signal, effectively filtering out non-linear noise artifacts caused by ambient lighting or complex motion.

	\textbf{(2) Phase Intervention}
	
	This intervention involves shifting the phase of the signal in both the video and signal domains. To avoid interference between the original physiological pulsation and the newly injected target waveform, we employ a `remove-then-add' strategy. Concretely, we refer to the method of `Falsifiability via Nulling Objective' and use the editor to create the null-pulse video $X_{nul}$. We then inject the transformed target signal into this null-pulse video. Let $\tau(\cdot)$ be the phase shift operator. The formula is:
	
	\vspace{-0.9em}
	\begin{equation}
		\begin{aligned}
			&X_\tau = G(X_{nul}, \tau(s_0)),\\
			&E(X_\tau) \approx \tau(s_0).
		\end{aligned}
	\end{equation}

	\textbf{(3) Frequency Intervention}
	This intervention modifies the signal frequency in both domains. The procedure is identical to phase intervention, embedding a frequency-transformed rPPG signal onto the null-pulse video $X_{nul}$. Let $\rho(\cdot)$ be the frequency scaling operator. The formula is:
	
	\vspace{-0.9em}
	\begin{equation}
		\begin{aligned}
			&X_\rho = G(X_{nul}, \rho(s_0)),\\
			&E(X_\rho) \approx \rho(s_0).
		\end{aligned}
	\end{equation}
	
	Phase and frequency interventions compel the model to achieve temporal disentanglement. The model must learn to recognize the waveform morphology of the signal, and this recognition ability should not depend on the signal's absolute temporal position or its fundamental frequency. This is crucial for the model to robustly handle individuals with different heart rates.

	\begin{figure*}[t]
		\centering
		\includegraphics[scale=0.47]{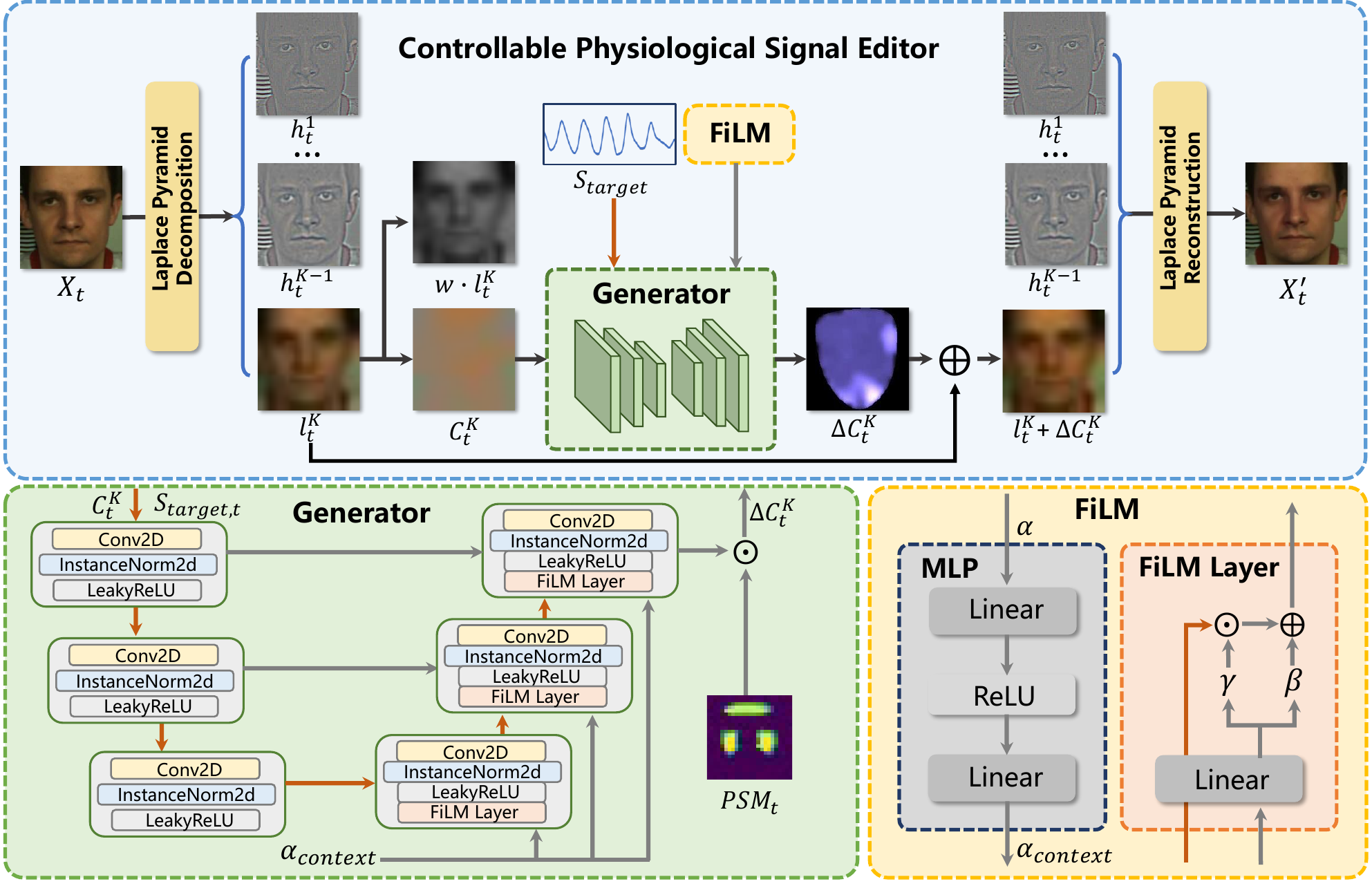}
		\vspace{-0.3em}
		\caption{\small{
				Framework of the Controllable Physiological Signal Editor. An input frame ($X_t$) is first separated into high-frequency layers ($\{h_t^k\}$) and a low-frequency base layer ($l_t^K$) using Laplacian Pyramid Decomposition. The low-frequency base is converted into a luminance-suppressed chrominance carrier ($C_t^K$), on which a U-Net based generator predicts a residual physiological perturbation. The target signal ($S_{target}$) provides temporal control over the sign and evolution of the perturbation, while the perturbation support map ($PSM_t$) provides spatial support by restricting the perturbation to reliable skin regions. The control parameter ($\alpha$) modulates the intervention strength through FiLM layers. The predicted chrominance perturbation ($\Delta C_t^K$) is then added back to the low-frequency layer, and the modified low-frequency layer is recombined with the unchanged high-frequency layers to reconstruct the edited frame ($X_t'$), preserving video fidelity.
			}
		}
		\label{fig:SignalEditor}
		\vspace{-0.3em}
	\end{figure*}  
	
	\vspace{0.3em}
	\noindent\textbf{Spatio-temporal Prior Objective.}\quad  
	In addition to the physical axioms governing the signal, we can further enhance the quality of the learned signal by imposing constraints derived from spatio-temporal priors of the rPPG signal.
	
	Prior research \cite{niu2017continuous} indicates that periodic changes in the chrominance channels of human skin directly correspond to the physiological signal caused by blood flow. Therefore, we add a forward constraint that maximizes the cross-correlation between the extracted signal $s_0$ and a robust low-frequency chrominance reference signal $C_{obs}$, ensuring that the learned signal can genuinely explain the observed phenomena in the video.

	In the spatial domain, we introduce a multi-region consistency constraint and a background suppression constraint. First, we ensure that the extracted signal $s_0$ is consistent with signals $c_k$ from different facial regions. Second, we enforce that the signal from background regions, $c_{bg}$, should have both its energy and its correlation with the signal $s_0$ approaching zero. Both $c_k$ and $c_{bg}$ are extracted by the same extractor from video clips with different masked regions set.

	\subsection{Interv-rPPG Framework}
	\label{sec:Interv-rPPG}
	
	To implement the PCP paradigm, we propose a self-supervised causal intervention framework, Interv-rPPG, consisting of an rPPG signal extractor and a signal editor.

	\vspace{0.3em}
	\noindent\textbf{PhysMambaFormer Extractor.} \quad 
	
	The rPPG extractor $E$ within our framework must be fast and accurate. We propose PhysMambaFormer, a hybrid architecture that combines the linear-complexity long-range modeling ability of State Space Models (SSMs)\cite{voelker2018improving} with the global consistency alignment of attention mechanisms.
	
	As illustrated in Fig.~\ref{fig:PhysMambaFormer}, given an input clip $X \in \mathbb{R}^{C \times T \times H \times W}$, we first apply temporal normalization to each pixel along the temporal axis to remove slow drift and normalize temporal variance. A lightweight 3D-CNN then extracts local spatio-temporal features, which are reshaped frame-wise and projected by a linear layer with dropout into latent tokens $Z_t^{proj} \in \mathbb{R}^{d}$. To suppress slow-varying nuisance factors while retaining pulse-related dynamics, we add a first-order temporal-difference prior in the latent space, i.e., $Z_t^{proj}\gets Z_t^{proj}+\left(Z_t^{proj}-Z_{t-1}^{proj}\right)$, with zero padding for the first frame. The resulting tokens are processed by 4 Gated Temporal State Space (GTSS) blocks. In each block, an input projection splits the token $Z_t^{proj}$ into a main branch $X_t^{core}$ and a gating branch $R_t$. The main branch first passes through a depth-wise temporal convolution and a SiLU activation, and is then fed into a selective state-space scan mechanism. Within this mechanism, the model dynamically projects and generates the core discretized transition matrix $\Delta A$, the input driving term $\Delta B$, and the output projection matrix $C$ based on the current input, while $D$ serves as a learned global skip connection parameter. Its core scan process involves an iterative update on a hidden state, where the state $h_t$ is updated by combining the previous state transformed by $\Delta A$ and the current input term $\Delta B$. The final scan output $y$ is modulated by $SiLU(R_t)$ from a parallel branch, projected back to the model dimension, and added to the module input $Z_t^{proj}$ via an outer residual connection. After GTSS blocks, a Global Attention Layer (GAL) captures clip-level consistency across all time steps, and a linear rPPG regression head outputs one scalar per frame. Therefore, for an input clip of length $T$, the predicted rPPG waveform also has length $T$. During inference, we further apply a fixed-coefficient FIR band-pass filter with passband 0.67--4.0 Hz (40--240 bpm) to suppress out-of-band noise.

	\noindent\textbf{Controllable Physiological Signal Editor.}  \quad 
	To achieve precise and controllable intervention on the rPPG signal while maintaining high video fidelity, we designed a controllable physiological signal editor based on the Laplacian pyramid. Its core principle is to decouple image content in the frequency domain and perform targeted modulation only on the low-frequency chrominance component that carries the physiological signal.
	
	As illustrated in Fig.~\ref{fig:SignalEditor}, first, each frame $X_t$ of the video is decomposed using a Laplacian pyramid \cite{wu2012eulerian} into $L(X_t)=\{h_t^1,...,h_t^{K-1},l_t^K\}$, where $\{h_t^k\}_{k=1}^{K-1}$ are the high-frequency layers containing texture and edge details, and $l_t^K$ is the low-frequency base layer containing color and illumination. Since the rPPG signal is manifested mainly through subtle color variations rather than luminance changes, we first suppress the luminance-dominant component in $l_t^K$ using a fixed standard luminance weighting vector $w$, thereby obtaining the luminance suppressed chrominance carrier $C_t^K$:
	
	\vspace{-0.9em}
	\begin{equation}
		\begin{aligned}
			C_t^K = l_t^K - (l_t^K \cdot w).
		\end{aligned}
	\end{equation}

	\begin{figure}[t]
		\centering
		\includegraphics[scale=0.26]{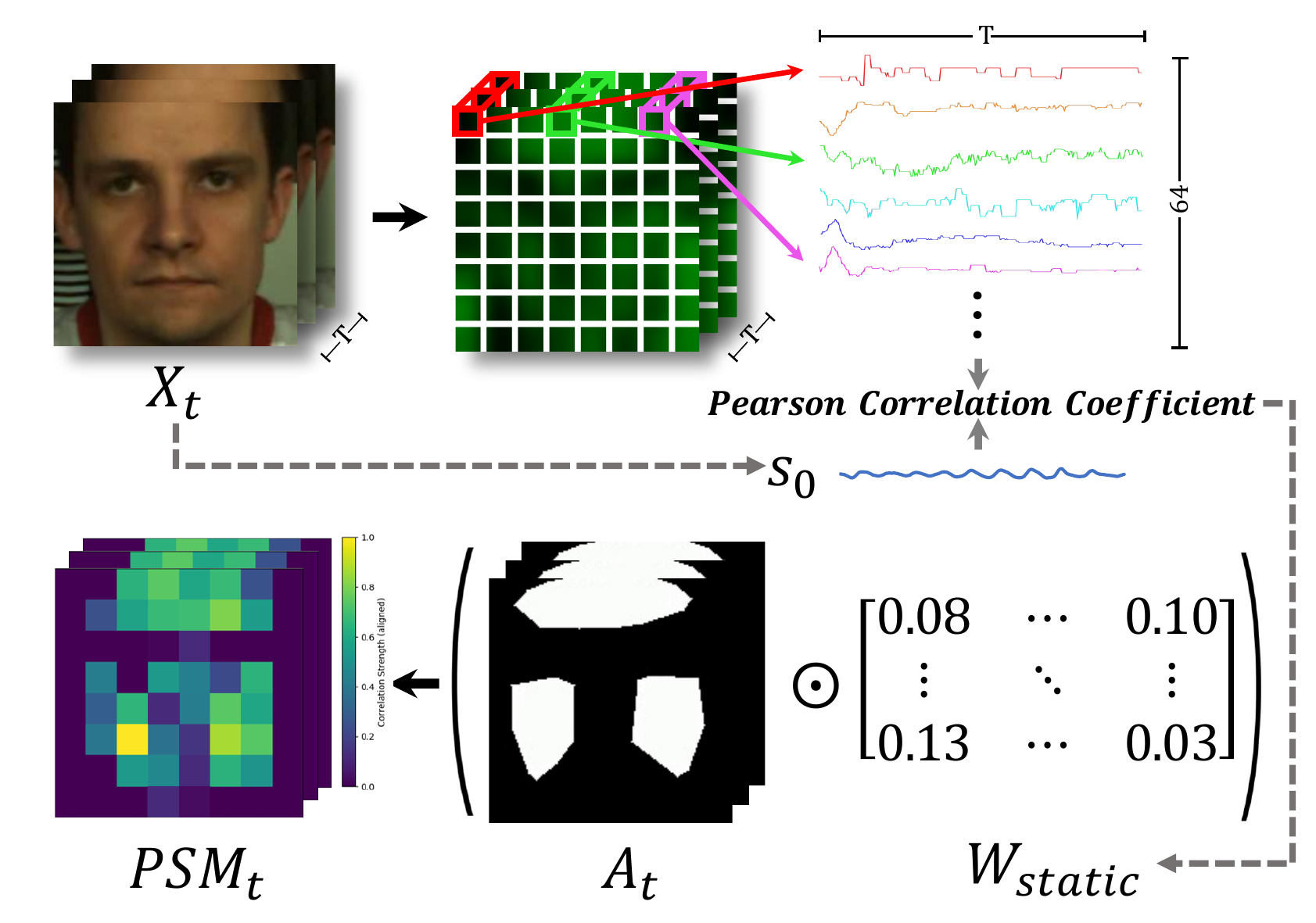}
		\vspace{0em}
		\caption{\small{
				Perturbation Support Map. For each frame ($X_t$), a Dynamic Anatomical Prior ($A_t$) is computed, which is a soft mask emphasizing reliable skin regions (forehead, cheeks) while suppressing eyes and mouth. Static Consensus Prior ($W_{static}$) is an $8\times8$ correlation map: the low-frequency chrominance green channel is partitioned into 64 $8\times8$ grid signals, with Pearson correlation to $s_0$ forming $W_{static}$. The final map $PSM_t$ is the element-wise product of $A_t$ and $W_{static}$.
			}
		}
		\label{fig:PSM}
		\vspace{-0.3em}
	\end{figure}

	The generator then predicts a low-frequency chrominance perturbation $\Delta C_t^K$ under joint temporal-spatial conditioning. Temporally, the target physiological signal $S_{target}$ specifies the sign, timing, and desired evolution of the perturbation. Spatially, the Perturbation Support Map $PSM$ determines where on the face the perturbation is allowed to appear. Specifically, $S_{target,t}$ is broadcast to the same spatial resolution as $C_t^K$, and the two are concatenated channel-wise as the input to the generator (U-Net architecture \cite{ronneberger2015u}). To enable flexible control over the intervention strength, a small MLP maps the user-specified control parameter $\alpha$ to Feature-wise Linear Modulation \cite{perez2018film} (FiLM) parameters, which are used to modulate decoder features. The decoder predicts a chrominance residual map, and the final perturbation is obtained by multiplying this output with the perturbation support map $PSM_t$:
	
	\vspace{-0.9em}
	\begin{equation}
		\begin{aligned}
			\Delta C_t^K = Decoder(C_t^K, S_{target,t}; \alpha)\odot PSM_t.
		\end{aligned}
	\end{equation}
	
	The edited frame is then reconstructed as:
	
	\vspace{-0.9em}
	\begin{equation}
		\begin{aligned}
			X_t^{'} = Reconstruct(\{h_t^1,...,h_t^{K-1},l_t^K+\Delta C_t^K\}).
		\end{aligned}
	\end{equation}
	
	Because the intervention is confined to the low-frequency chrominance domain, the high-frequency identity information and most of the original appearance details are preserved.
	
	In addition, the process of calculating $PSM$ is shown in Fig.~\ref{fig:PSM}. To ensure physiological rationality and temporal stability of perturbations, we first build a frame-wise facial dynamic anatomical prior $A_t$, anchored to high-SNR skin regions (forehead, cheeks) and suppressing non-physiological regions (eyes, mouth). We then compute a static consensus prior $W_{\text{static}}$ from the low-frequency chrominance observations. Specifically, we take the green channel of the low-frequency chrominance sequence, uniformly partition it into an $8\times 8$ grid, and average the pixels within each grid cell along the spatial dimensions to obtain 64 local temporal signals. The Pearson correlation coefficient between each local signal and the current source hypothesis $s_0$ is then calculated, and the resulting $8\times 8$ correlation map is used as the static consensus prior $W_{\text{static}}$. Finally, the perturbation support map is defined as: $PSM_t = A_t \odot W_{\text{static}}$.

	\noindent\textbf{Losses for Interv-rPPG.} \quad 
	To implement three objectives from the PCP paradigm in Section~\ref{sec:PCP}, we construct a composite loss function with three main components.

	\textit{Falsifiability via Nulling Loss}: This loss implements the `Falsifiability via Nulling' objective. The extractor $E$ outputs a hypothesis $s_0$, and the editor $G$ generates a null-pulse video $X_{nul}$ by injecting the anti-signal $-s_0$. The extractor then estimates the residual signal $s_{nul} = E(X_{nul})$. The loss $\mathcal{L}_{nul}$ minimizes the energy of $s_{nul}$ in the physiological frequency band:

	\vspace{-0.9em}
	\begin{equation}
		\begin{aligned}
			\mathcal{L}_{nul} = \frac{1}{T} \sum_{t=1}^{T} \left(BP(s_{nul})_t \right)^2,
		\end{aligned}
	\end{equation}
	where $T$ is the signal length and $BP(\cdot)$ is a fixed FIR filter for 0.67–4.0 Hz (40–240 bpm). This loss ensures $s_0$ is physically cancellable by chrominance intervention, thereby rejecting non-cancellable artifacts like motion.

	\textit{Axiomatic Equivariance Loss}: This loss executes the `Axiomatic Equivariance' objective, ensuring the learned representation adheres to signal physics. Based on the signal hypothesis $s_0$ proposed by the extractor $E$, we use the editor $G$ to separately modify the amplitude, phase, and frequency of the signal in the video domain, yielding the corresponding modified videos $X_\alpha$, $X_\tau$, and $X_\rho$. We anticipate that the output of the extractor $E$ should also exhibit corresponding physical changes. Thus, we define the equivariant axiomatic loss as follows:
	
	\vspace{-0.9em}
	\begin{equation}
		\begin{aligned}
			\mathcal{L}_{equ}&=\vert BP(E(X_\alpha))-\alpha\cdot BP(s_0) \vert \\& + \vert BP(E(X_\tau))-\tau({BP(s}_0)) \vert \\& + \vert BP(E(X_\rho))-\rho(BP(s_0))\vert,
		\end{aligned}
	\end{equation}
	where $\vert \cdot \vert$ denotes the sample-wise absolute difference on the temporal waveform, followed by averaging over time.

			\begin{algorithm}[h!]
		\caption{Stage-III Self-Supervised Training Process}
		\label{alg:pcp_train}
		\begin{algorithmic}[1]
			\Require Randomly initialized extractor $E(\theta)$, Stage-II pre-trained frozen editor $G$, Video $X$, Observed chrominance $C_{\text{obs}}$, Warm-up epochs $T_{\text{warmup}}$, Nulling search range $[\alpha_{\text{min}}, \alpha_{\text{max}}]$
			
			\State // --- 1. Get extractor’s hypothesis ---
			\If{epoch $< T_{\text{warmup}}$}
			\State $s_{0} \leftarrow \text{POS}(X)$ \Comment{Use proxy signal}
			\Else
			\State $s_0 \leftarrow E(X)$ \Comment{Extract with $E$}
			\EndIf
			
			\State // --- 2. Falsifiability via Nulling ---
			\State $E_{\text{best}} \leftarrow \infty$, $\alpha^* \leftarrow 0$
			\For{$\alpha^*_{cur} \in [\alpha_{\text{min}}, \alpha_{\text{max}}]$} \Comment{Search for best $\alpha^*$}
			\State $X_{\alpha^*} \leftarrow G(X, \alpha^*_{cur} \cdot s_{0})$
			\State $E_{\text{current}} \leftarrow \left\| \text{bandpass}(E(X_{\alpha^*})) \right\|^2$ \Comment{Energy of video data}
			\If{$E_{\text{current}} < E_{\text{best}}$}
			\State $E_{\text{best}} \leftarrow E_{\text{current}}$, $\alpha^* \leftarrow \alpha^*_{cur}$
			\EndIf
			\EndFor
			\State $X_{\text{nul}} \leftarrow G(X, \alpha^* \cdot s_{\text{0}})$ \Comment{Generate pulse-free video}
			\State $\mathcal{L}_{\text{nul}} \leftarrow \left\| \text{bandpass}(E(X_{\text{nul}})) \right\|^2$ 
			
			\State // --- 3. Axiomatic Equivariance ---
			\State $T_s \leftarrow \text{Transform}(\text{amplitude, phase, frequency})$
			\State $s_{\text{target}} \leftarrow T_s(s_{\text{0}})$
			\If{$T_s = \text{amplitude}$}
			\State $x^{'} \leftarrow G(X, s_{\text{target}})$
			\Else
			\State $x^{'} \leftarrow G(X_{\text{nul}}, s_{\text{target}})$ \Comment{Remove-then-add}
			\EndIf
			
			\State $s_{\text{pred}} \leftarrow E(X^{'})$
			\State $\mathcal{L}_{\text{equ}} \leftarrow \vert s_{\text{pred}} - s_{\text{target}} \vert$
			
			\State // --- 4. Spatio-temporal Priors ---
			\State $\mathcal{L}_{\text{ST}} \leftarrow \text{Loss}(s_0, C_{\text{obs}}, c_{\text{bg}}, c_k) + \mathcal{L}_{\text{wave}}$
			
			\State // --- 5. Total Loss and Update ---
			\State $\mathcal{L} \leftarrow \mathcal{L}_{\text{nul}} + \mathcal{L}_{\text{equ}} + \mathcal{L}_{\text{ST}}$
			\State Update $\theta$ (Extractor $E$) using $\nabla_{\theta} \mathcal{L}$
		\end{algorithmic}
	\end{algorithm}

	\textit{Spatio-temporal Prior Loss}: This loss denoted as $\mathcal{L}_{ST}$ incorporates physiological and signal-level priors to enhance the quality of $s_0$. It includes several key components: First, the forward constraint encourages $s_0$ to be highly correlated with the robust reference signal $C_{obs}$. Specifically, after computing the static consensus prior $W_{\text{static}}$, we select the top-$K$ most reliable local facial cells within the anatomically reliable facial support defined by $A_t$. We then extract their low-frequency chrominance time series directly from the visual input. The final reference signal $C_{obs}$ is obtained as the point-wise median of these local signals, and ultimately obtained through bandpass filtering operation. This ensures $E$'s hypothesis can genuinely explain the phenomena in the video. Second, the multi-region consistency constraint encourages $s_0$ to be consistent with local signals $c_k$ extracted by the same $E$ under different local facial masks corresponding to these top-$K$ cells. Third, the background suppression constraint enforces low correlation and low energy for the background signal $c_{bg}$, which is extracted by the same $E$ from the background region. Finally, the signal fidelity constraint $\mathcal{L}_{wave}$ imposes a prior on the waveform of $s_0$ itself, which includes constraining spectral concentration (i.e., low spectral entropy) and temporal consistency of the signal across windows. We compute the Power Spectral Density ($PSD(f)$) of the signal within the frequency band $[f_{l}, f_{h}]$ to constrain the energy concentration of the signal in the heart rate band, while utilizing the Jensen-Shannon ($JS$) divergence to enforce the consistency between the spectral distributions $PSD_1(f)$ and $PSD_2(f)$ of the front and rear segments of the signal.
	
	\vspace{-0.9em}
	\begin{equation}
		\begin{aligned}
			\mathcal{L}_{wave} &= - \sum_{f \in [f_l, f_h]} PSD(f) \cdot \log(PSD(f) + \epsilon) \\&+ JS (PSD_1(f), PSD_2(f)),
		\end{aligned}
	\end{equation}	
	
	\begin{equation}
		\begin{aligned}
			\mathcal{L}_{ST} &= 1 - \vert \mathcal{P}(s_0, C_{obs}) \vert + \frac{1}{K}\sum_{k=1}^{K} \left(1.0 - \vert \mathcal{P}(s_0, c_k) \vert \right)  \\
			&\quad + \vert \mathcal{P}(s_0, c_{bg}) \vert + \mathcal{L}_{wave},
		\end{aligned}
	\end{equation}
	where $\mathcal{P}(\cdot, \cdot)$ calculates the pearson correlation coefficient between two signals.

	\textit{Overall Loss}: The overall loss function for Interv-rPPG is the combination of the three losses, which can be expressed as:
	
	\vspace{-0.9em}
	\begin{equation}
		\begin{aligned}
			\mathcal{L} = \mathcal{L}_{nul} + \mathcal{L}_{equ} + \mathcal{L}_{ST}.
		\end{aligned}
	\end{equation}

	\noindent\textbf{Three-stage training protocol.} \quad The complete pipeline contains three explicitly separated training stages. \textbf{Stage-I} trains a supervised extractor $E^{*}$ from labeled videos. It is used only to pre-train the editor $G$ and to provide the supervised baseline reported as PhysMambaFormer*. \textbf{Stage-II} freezes $E^{*}$ and trains the editor $G$ to perform nulling, amplitude, phase, and frequency interventions under the supervision of $s_{gt}$. \textbf{Stage-III} discards $E^{*}$ and introduces a new extractor $E$ with random initialization. $E$ is optimized with frozen $G$ and unlabeled videos under the PCP objectives.
	
	Specifically, during Stage-II, the editor $G$ is first trained to ensure its capability to accurately modify rPPG signals in videos: a video reconstruction loss is defined to maintain the visual consistency between the reconstructed video and the original one, while a Falsifiability via Nulling Loss and an Axiomatic Equivariance Loss are employed to enable the editor to master precise rPPG signal editing; in this stage, the ground-truth (GT) rPPG signal $s_{gt}$ inherent to the video is used as input, and $E^{*}$ is utilized to verify whether the editor has accurately achieved rPPG signal editing. The total loss of the editor $\mathcal{L}_{G}$ is as follows: 
	
	\vspace{-0.9em}
	\begin{equation}
		\begin{aligned}
			\mathcal{L}_{G} &= \frac{1}{T} \sum_{t=1}^{T} \left( X_t - X_t^{'} \right)^2 + \\
			&\quad \frac{1}{T} \sum_{t=1}^{T} \left(BP(E^{*}(G(X,-s_{gt})))_t \right)^2 + \\
			&\quad \vert BP(E^{*}(X_\alpha))-\alpha \cdot BP(s_{gt})\vert + \\
			&\quad \vert BP(E^{*}(X_\tau))-\tau(BP(s_{gt})) \vert + \\
			&\quad \vert BP(E^{*}(X_\rho))-\rho(BP(s_{gt})) \vert.
		\end{aligned}
	\end{equation}
	
	For completeness, we also study a fully unlabeled Stage-II variant in which the editor-training objective above replaces $E^{*}$ with POS and replaces $s_{gt}$ with the POS signal extracted from the original video. The corresponding Stage-III model is denoted as Interv-rPPG(unlabeled) in the experiments.
	
	\noindent\textbf{Self-Supervised Training Process.} \quad 
	The Stage-III PCP training relies on a trainable extractor $E$ with random initialization and a pre-trained, frozen editor $G$. The self-supervised training process begins with a $T_{warmup}$ phase. During this initial stage, the signal $s_0$ used to drive the interventions is not taken from the extractor. Instead, we use the traditional POS algorithm to calculate the proxy signal from the video data as $s_0$. This ensures that the interventions are physically-grounded from the very beginning. After the warm-up period, $s_0$ is taken from the extractor's own hypothesis $E(X)$.
	
	In the implementation of `Falsifiability via Nulling' objective. We do not simply assume that injecting $-s_0$ will nullify the signal. Instead, we perform a search in the data domain for an optimal scalar $\alpha^* \in [\alpha_{min}, \alpha_{max}]$ that actually minimizes the energy of the edited video $G(X, \alpha^* \cdot s_{\text{0}})$. This search yields a pulse-free video $X_{nul}$. The extractor is then trained via $\mathcal{L}_{nul}$ to output near-zero energy when observing this $X_{nul}$. In practice, we adopt a coarse-to-fine search that first scans 5 uniformly sampled $\alpha$ values and then performs a 3-step local refinement around the best candidate.
	
	In `Axiomatic Equivariance' phase, $E$ is trained to learn a consistent mapping that adheres to the physical transformations (amplitude, phase, frequency) performed by $G$, while also being anchored by external physical priors loss $\mathcal{L}_{ST}$. The complete training logic is summarized in Algorithm~\ref{alg:pcp_train}.

	\vspace{-0.3em}
	\section{Experimental Results}
	\vspace{-0.2em}
	\label{sec:experiemnts}

	\begin{table*}[]
		\centering
		\caption{Intra-dataset HR estimation results on the VIPL-HR, MMPD, PURE and UBFC-rPPG datasets.} 
		\label{tab:intra}
		\resizebox{\textwidth}{!}{
			\begin{threeparttable} 
				\begin{tabular}{cccccccccccccc}
					\hline
					\multirow{2}{*}{Type} & \multirow{2}{*}{Method} & \multicolumn{3}{c}{VIPL-HR} & \multicolumn{3}{c}{MMPD} & \multicolumn{3}{c}{PURE} & \multicolumn{3}{c}{UBFC-rPPG} \\ \cline{3-14} 
					& & \thead{MAE\\ (bpm)} & \thead{RMSE\\ (bpm)} & r & \thead{MAE\\ (bpm)} & \thead{RMSE\\ (bpm)} & r & \thead{MAE\\ (bpm)} & \thead{RMSE\\ (bpm)} & r & \thead{MAE\\ (bpm)} & \thead{RMSE\\ (bpm)} & r \\ \hline
					\multirow{3}{*}{Traditional} & POS \cite{wang2016algorithmic} & 11.50 & 17.20 & 0.30 & 12.36 & 17.71 & 0.18 & - & - & - & 4.05 & 8.75 & 0.78 \\
					& GREEN \cite{verkruysse2008remote} & 15.90 & 21.00 & 0.11 & 21.68 & 27.69 & -0.01 & - & - & - & 7.50 & 14.41 & 0.62 \\
					& CHROM \cite{de2013robust} & 11.40 & 16.90 & 0.28 & 13.66 & 18.76 & 0.08 & 2.07 & 9.92 & \textbf{0.99} & 2.37 & 4.91 & 0.89 \\ \hline
					\multirow{5}{*}{Supervised} & PhysNet \cite{yu2019remote1} & 10.80 & 14.80 & 0.20 & \textbf{4.81} & 11.83 & 0.60 & 2.10 & 2.60 & \textbf{0.99} & - & - & - \\
					& RhythmNet \cite{niu2019rhythmnet} & 5.30 & 8.14 & 0.76 & - & - & - & - & - & - & 1.19 & 2.10 & {\ul0.98} \\
					& PhysFormer \cite{yu2022physformer} & 4.97 & 7.79 & 0.78 & 11.99 & 18.41 & 0.18 & - & - & - & - & - & - \\
					& Dual-GAN \cite{lu2021dual} & 4.93 & 7.68 & {\ul0.81} & - & - & - & 0.82 & 1.31 & \textbf{0.99} & 0.44 & 0.67 & \textbf{0.99} \\
					& rPPG-MAE \cite{liu2024rppg} & {\ul4.52} & {\ul7.49} & {\ul0.81} & - & - & - & {\ul0.40} & {\ul0.92} & \textbf{0.99} & \textbf{0.17} & \textbf{0.21} & \textbf{0.99} \\
					& PhysMambaFormer* & 4.61 & 7.69 & 0.80 & 6.88 & 10.42 & 
					0.63 & 0.85 & 1.30 & \textbf{0.99} & {\ul0.41} & {\ul0.62} & \textbf{0.99} \\ \hline
					Semi-supervised & Semi-rPPG \cite{wu2025semi} & 6.06 & 10.99 & 0.57 & - & - & - & - & - & - & - & - & - \\ \hline
					\multirow{5}{*}{Self-supervised} & Gideon21 \cite{gideon2021way} & 9.80 & 15.48 & 0.38 & - & - & - & 2.30 & 2.90 & \textbf{0.99} & 1.85 & 4.28 & 0.93 \\
					& Contrast-Phys+ \cite{sun2024contrast} & - & - & - & - & - & - & 1.00 & 1.40 & \textbf{0.99} & 0.64 & 1.00 & \textbf{0.99} \\
					& RS-rPPG \cite{savic2024rs} & 5.98 & 10.50 & 0.56 & - & - & - & \textbf{0.29} & \textbf{0.59} & \textbf{0.99} & - & - & - \\
					& Interv-rPPG(unlabeled) & 4.73 & 7.61 & 0.80 & 6.74 & {\ul 10.01} & {\ul 0.64} & 1.04 & 1.66 & 0.99 & 0.69 & 1.20 & 0.98 \\
					& \textbf{Interv-rPPG(Ours)} & \textbf{4.47} & \textbf{7.26} & \textbf{0.83} & {\ul 6.31} & \textbf{9.27} & \textbf{0.69} & 0.92 & 1.52 & \textbf{0.99} & 0.56 & 1.01 & \textbf{0.99} \\ \hline
				\end{tabular}
				\begin{tablenotes}   
					\footnotesize            
					\item Boldfaced text indicates the best results, while underlined text denotes the second-best results.
					\item PhysMambaFormer* denotes a supervised extractor trained on Stage-I,  Interv-rPPG(unlabeled) denotes the variant in which Stage-II editor training uses POS in place of $E^{*}$ and $s_{gt}$.
				\end{tablenotes}
			\end{threeparttable} 
		}
	\end{table*}

	\subsection{Experimental setups and datasets}
	\label{sec:dataset} 
	\vspace{-0.2em}

	\noindent\textbf{Datasets.} \quad 
	\textcolor{red}We conducted experiments on four datasets, which are categorized into two types of scenarios: `clean' datasets (UBFC-rPPG \cite{bobbia2019unsupervised}, PURE \cite{stricker2014non}) characterized by a neat background, small subject movement amplitude, and relatively stable lighting; and more challenging datasets (VIPL-HR \cite{niu2018vipl}, MMPD \cite{tang2023mmpd}) featuring larger motion, more diverse devices, or more complex lighting conditions. Details of each dataset are as follows:
	
	\textit{UBFC-rPPG Dataset} \cite{bobbia2019unsupervised}: It consists of 42 uncompressed videos of 42 subjects, with a resolution of 640×480 pixels and a frame rate of 30 fps. This dataset synchronously records the photoplethysmography (PPG) waveforms, heart rate (HR) values, and blood oxygen saturation indices of the subjects, covering scenarios such as minor head movements and slow lighting changes.

	\textit{PURE Dataset} \cite{stricker2014non}: It contains 60 videos of 10 participants, each with a duration of 1 minute, a resolution of 640×480 pixels, and a frame rate of 30 fps. The dataset covers 6 scenarios: static state, speaking, slow translation, fast translation, small-angle rotation, and medium-angle rotation. The participants' heart rates exhibit a large fluctuation range, ranging from 42 to 148 bpm.

	\textit{VIPL-HR Dataset} \cite{niu2018vipl}: It includes 2,378 visible light (VIS) videos and 752 near-infrared (NIR) videos, involving 107 subjects. The videos are collected using 3 different devices and cover 9 complex scenarios (including drastic changes in light intensity and color temperature, rapid head movements, etc.), aiming to simulate various interference factors in real-world environments.
	
	\textit{MMPD Dataset} \cite{tang2023mmpd}: It is a mobile-phone rPPG dataset comprising approximately 11 hours of recordings from 33 subjects, with diverse skin tones, four lighting conditions, and four activity conditions including stationary, head rotation, talking, and walking. These characteristics make it a suitable benchmark for evaluating robustness under realistic motion and illumination variations.
	
	For both Stage-I supervised training and Stage-III self-supervised training, we use identical dataset partitions within each dataset. In intra-dataset evaluation, PURE uses the first 36 samples for training and the remaining 24 for testing; UBFC-rPPG uses the first 30 samples for training and the remaining 12 for testing; VIPL-HR follows the subject-exclusive 5-fold cross-validation protocol; and MMPD is split into training/validation/testing subsets with a 7:1:2 ratio. In cross-dataset evaluation, models are separately trained on PURE, UBFC-rPPG, and PURE+UBFC-rPPG, and the source-domain training pool is further divided into 80\% training and 20\% validation for model selection. The held-out target-domain evaluations are then performed on PURE, UBFC-rPPG, and VIPL-HR without using target-domain labels. Stage-II always uses only the labeled training split of the current source domain(s).

	\noindent\textbf{Implementation Details.} \quad 
	In the data preprocessing stage, we adopted the Mediapipe \cite{lugaresi2019mediapipe} model to perform 468 facial landmark detection on all frames of each video. Firstly, we calculated the bounding rectangle of the facial region enclosed by the 468 landmarks in each frame. Subsequently, we computed the bounding square of all frame-wise bounding rectangles for each video to ensure the facial region was fully contained within the video frame. Finally, using this bounding square as the reference, we cropped each frame of the video and resized the cropped image to 192×192 pixels. In addition, we generated a facial mask sequence corresponding to each video by setting the facial region to 1 and other regions to 0. Considering the differences in the sampling rates of PPG signals recorded by different datasets, we resampled the length of the PPG signal corresponding to each video to match the total number of frames of the video. To adapt to the model input format, we cropped the video, its corresponding PPG signal, and the facial mask sequence into several non-overlapping clips with a length of 256. The PhysMambaFormer outputs a predicted rPPG signal with an output sequence length of 256 per clip, and the sampling rate of it is consistent with the input video frame rate (30 fps for all datasets used in this work).
	
	The experiments were conducted based on the PyTorch framework and executed on a single NVIDIA GeForce RTX 4090 GPU with 24GB video memory. In the self-supervised learning phase, the batch size was set to 4, the number of training epochs was 100, and the AdamW \cite{loshchilov2017decoupled} optimizer with a learning rate of 1e-4 was adopted, combined with the CosineAnnealingLR scheduler for dynamic learning rate adjustment. The stochastic intervention parameters used in training were set as follows: $\alpha \in [0.0, 2.0]$, $\tau \in [-12, 12]$ frames, and $\rho \in [0.8, 2.0]$. These intervention parameters are resampled at every batch. Stage-I trains the supervised extractor $E^{*}$ with labeled clips, Stage-II trains the editor $G$ for 60 epochs with frozen $E^{*}$, and Stage-III trains a randomly initialized extractor $E$ for 100 epochs with the frozen $G$. In Stage-III, the warm-up duration is 5 epochs.
	
	During the supervised fine-tuning phase of the model, the batch size was set to 8, the number of training epochs was 50, and the AdamW optimizer with a learning rate of 1e-4 was also used, paired with the ReduceLROnPlateau scheduler to realize dynamic learning rate adjustment.

	\noindent\textbf{Evaluation Metrics.} \quad 
	In the rPPG prediction task, following prior work \cite{yu2023physformer++, yu2019remote2, yu2020autohr}, we selected Mean Absolute Error (MAE), Root Mean Square Error (RMSE), and Pearson Correlation Coefficient (r) as the core evaluation metrics to quantify the consistency between the predicted heart rate and the ground-truth heart rate. Among these metrics, the lower the values of MAE and RMSE, the smaller the heart rate prediction error and the higher the prediction accuracy of the model; the closer the value of the Pearson Correlation Coefficient is to 1, the better the linear fitting effect between the predicted heart rate and the ground-truth heart rate. In the video rPPG signal editing task, we adopted Peak Signal-to-Noise Ratio (PSNR), Structural Similarity Index Measure (SSIM), and Learned Perceptual Image Patch Similarity (LPIPS) \cite{zhang2018unreasonable} to evaluate the quality of the edited video frames. Specifically, the higher the values of PSNR and SSIM, the higher the similarity between the edited video frames and the original ones, indicating a more desirable editing effect; on the contrary, the lower the value of LPIPS, the higher the perceptual similarity between the two frames of images, meaning the editing process causes less damage to the video quality.

		\begin{table}[t]
		\centering
		\caption{Cross-dataset HR estimation (training on UBFC-rPPG and testing on PURE).} 
		\label{tab:cross1}
		\begin{threeparttable} 
			\begin{tabular}{cccc}
				\hline
				Method                      & \thead{MAE\\(bpm)} & \thead{RMSE\\(bpm)}  & r             \\ \hline
				TS-CAN \cite{liu2020multi}          & 3.69     & 13.80   & 0.82    \\
				PhysFormer \cite{yu2022physformer}  & 12.92    & 24.36   & 0.47   \\
				DeepPhys \cite{chen2018deepphys}    & 5.54     & 18.51    & 0.66    \\
				Semi-rPPG \cite{wu2025semi}         & 2.35 & 7.49          & {\ul 0.95} \\
				RErPPG-Net \cite{hsieh2022augmentation} & 4.36          &  6.69    & 0.60     \\
				PhysMambaFormer* & {\ul 2.23}     & {\ul 5.41}    & {\ul0.95}     \\
				Interv-rPPG(unlabeled) & 2.31 & 5.74 & {\ul0.95} \\
				\textbf{Interv-rPPG(Ours)} & \textbf{2.08} & \textbf{5.29} & \textbf{0.96} \\ \hline
			\end{tabular}
			\begin{tablenotes}   
				\footnotesize            
				\item Boldfaced text indicates the best results, while underlined text denotes the second-best results.
			\end{tablenotes}
		\end{threeparttable} 
	\end{table}

	\subsection{Results of Remote Heart Rate Measurement}
	
	\noindent\textbf{Intra-Dataset Evaluation.} \quad   
	We conducted intra-dataset evaluation on Interv-rPPG across four datasets, namely UBFC-rPPG, PURE, VIPL-HR, and MMPD, with the experimental results presented in Table~\ref{tab:intra}. To verify the performance superiority of Interv-rPPG, we selected representative methods from traditional approaches \cite{wang2016algorithmic, verkruysse2008remote, de2013robust}, supervised learning \cite{yu2019remote1, niu2019rhythmnet, yu2022physformer, lu2021dual, liu2024rppg}, semi-supervised learning \cite{wu2025semi}, and unsupervised learning \cite{gideon2021way, sun2024contrast, savic2024rs} for comparative experiments. On the VIPL-HR dataset, which features significant environmental interference, Interv-rPPG achieved the optimal performance, even outperforming supervised methods. A similar trend is observed on MMPD, where Interv-rPPG improves over the Stage-I supervised extractor PhysMambaFormer* from 6.88 MAE to 6.31 MAE. These two challenging benchmarks jointly support our main claim that the Stage-III self-supervised extractor learns stronger robustness to motion and illumination than the Stage-I supervised extractor $E^{*}$.
	
	\begin{table}[t]
		\centering
		\caption{Cross-dataset HR estimation (training on PURE and testing on UBFC-rPPG).} 
		\label{tab:cross2}
		\begin{threeparttable} 
			\begin{tabular}{cccc}
				\hline
				Method                 & \thead{MAE\\ (bpm)}  & \thead{RMSE\\ (bpm)}  & r             \\ \hline
				CHROM \cite{de2013robust}        & 3.10         & 6.84                & 0.93                \\
				PulseGAN \cite{song2021pulsegan} & 2.09         & 4.42                & {\ul 0.97}           \\
				Dual-GAN \cite{lu2021dual}       & \textbf{0.74}   & \textbf{1.02}       & \textbf{0.99}       \\
				rPPG-MAE \cite{liu2024rppg}      & {\ul1.28}         & {\ul 2.75}                & -                   \\
				RErPPG-Net \cite{hsieh2022augmentation} & 2.23         & 4.66                & 0.78                \\
				PhysMambaFormer* & 1.82     & 3.85    & 0.96     \\
				Interv-rPPG(unlabeled) & 1.89 & 3.90 & 0.96 \\
				\textbf{Interv-rPPG(Ours)} & 1.67 & 3.36          & {\ul 0.97} \\ \hline
			\end{tabular}
			\begin{tablenotes}   
				\footnotesize            
				\item Boldfaced text indicates the best results, while underlined text denotes the second-best results.
			\end{tablenotes}
		\end{threeparttable} 
	\end{table}

	On the PURE and UBFC-rPPG datasets, Interv-rPPG remains competitive but is slightly below PhysMambaFormer*, which is consistent with the clean-domain characteristic: when the native signal-to-noise ratio is already high, the editor-driven chrominance intervention can introduce a small amount of residual perturbation instead of removing dominant nuisance factors. This also helps explain the gap to RS-rPPG on PURE. RS-rPPG \cite{savic2024rs} relies on spatial-temporal maps, ROI averaging, and multiple prior-based contrastive constraints, which suppress nuisance factors without explicitly rewriting the video chrominance. By contrast, PCP intentionally trades a small amount of clean-domain fidelity for a much stronger intervention-based invariance on challenging data. Additionally, the fully unlabeled variant Interv-rPPG(unlabeled) is consistently slightly below the full model but remains strong on VIPL-HR and MMPD, indicating that the Stage-II labeled editor verification improves editing precision rather than defining the learned Stage-III representation.

	\begin{table}[t]
		\centering
		\caption{Cross-dataset HR estimation (training on UBFC-rPPG+PURE and testing on VIPL-HR).} 
		\label{tab:cross3}
		\begin{threeparttable} 
			\begin{tabular}{cccc}
				\hline
				Method                       & \thead{MAE\\(bpm)}  & \thead{RMSE\\(bpm)}  & r         \\ \hline
				DG-rPPGNet \cite{chung2022domain}   & 18.23   & 18.81   & - \\
				RErPPG-Net \cite{hsieh2022augmentation} & 25.40     & 31.14          & 0.15 \\
				PhysMambaFormer* & 13.85 & 17.58 & 0.63 \\
				Interv-rPPG(unlabeled) & {\ul 11.32} & {\ul 16.28} & {\ul 0.66} \\
				\textbf{Interv-rPPG(Ours)} & \textbf{10.53} & \textbf{15.52} & \textbf{0.68} \\ \hline
			\end{tabular}
			\begin{tablenotes}   
				\footnotesize            
				\item Boldfaced text indicates the best results, while underlined text denotes the second-best results.
			\end{tablenotes}
		\end{threeparttable} 
	\end{table}

	\noindent\textbf{Cross-Dataset Evaluation.} \quad   
	To evaluate the domain generalization ability of Interv-rPPG under zero-shot setting, which is a key indicator to measure whether the model better resists the `correlation trap' and learns the physical properties of rPPG signals, we conducted a series of rigorous cross-dataset tests. More specifically, we separately trained on UBFC-rPPG, PURE, and UBFC-rPPG+PURE, and directly evaluated on unseen target domains without using any target-domain samples for training or model selection.
	
	As shown in Table~\ref{tab:cross1}, when the model was trained on UBFC-rPPG and tested on PURE, our Interv-rPPG achieved a MAE of 2.08, outperforming all comparative methods. This result indicates that the model possesses excellent transferability between `clean' domains with similar yet distinct characteristics. Although the fully unlabeled variant performs slightly worse than the full model, its performance is still adequate, indicating that the Stage-II labeled verification primarily improves intervention fidelity rather than serving as hidden supervision in Stage-III.
	
		\begin{figure}[t]
		\centering
		\includegraphics[width=\linewidth]{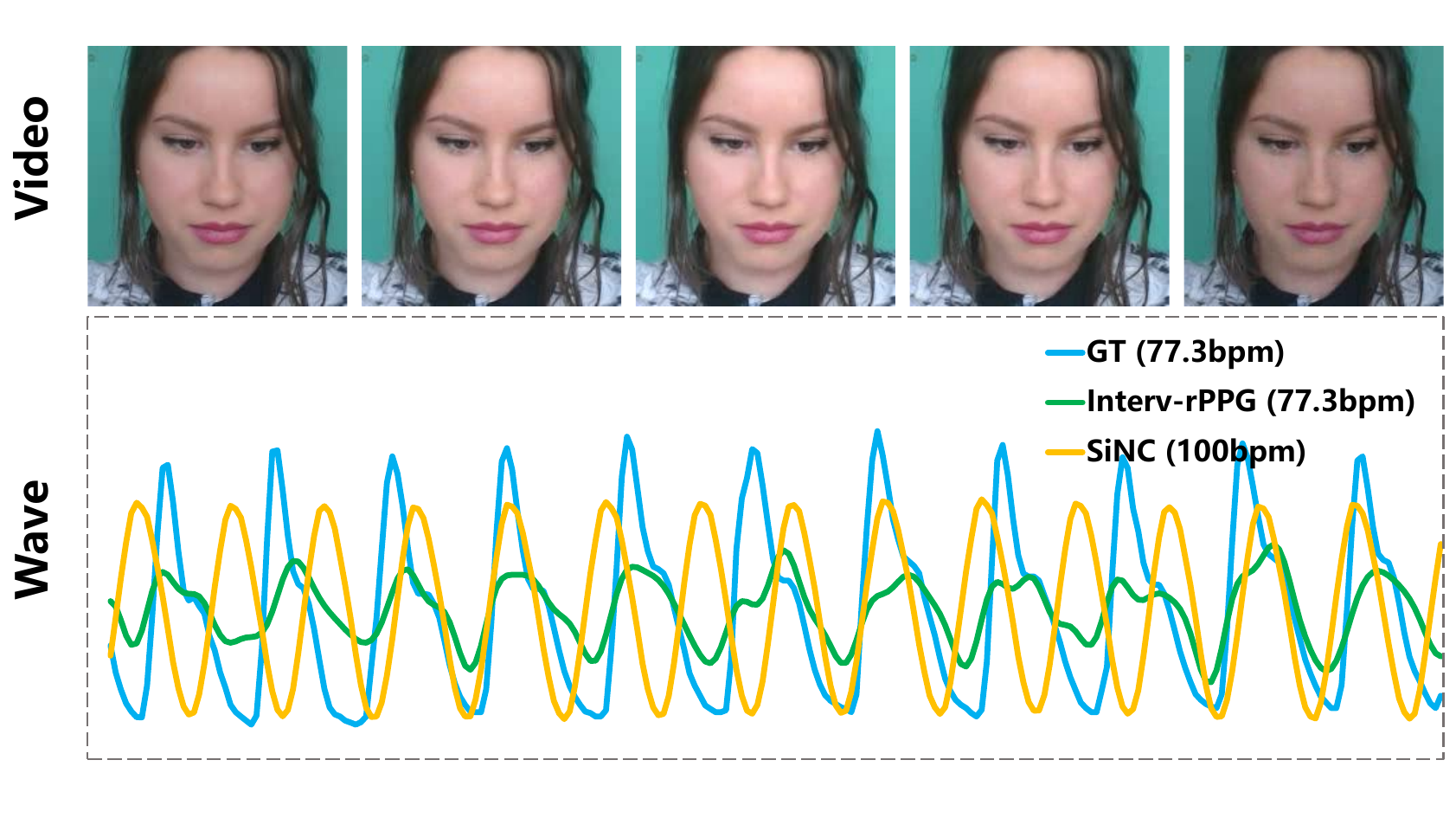}
		\caption{Visualization of rPPG signals extracted under periodic illumination perturbation. The first row shows the video clips with synthetic periodic illumination flicker (illumination frequency: 100 bpm). The second row presents the line chart of rPPG signals extracted by the existing SSL method SiNC and  Interv-rPPG(Ours) of the video after illumination perturbation. It can be clearly observed that the rPPG signal extracted by SiNC is affected by the illumination and its frequency is distorted to 100 bpm, while the frequency of the rPPG signal extracted by our Interv-rPPG method remains consistent with the original GT signal frequency, which further confirms the strong robustness of Interv-rPPG against periodic illumination interference.}
		\label{fig:illumination_visualization}
	\end{figure}

	In the reverse testing scenario presented in Table~\ref{tab:cross2} (trained on PURE and tested on UBFC-rPPG), Interv-rPPG exhibited a more remarkable performance advantage, achieving an MAE as low as 1.67. This result not only surpasses other self-supervised methods \cite{hsieh2022augmentation} but also is almost close to that of the supervised method rPPG-MAE \cite{liu2024rppg} on this task. This demonstrates that the physiological representations learned through the PCP paradigm are highly generalizable.

	Aiming to verify whether the model can generalize the knowledge learned from `clean' data to `in-the-wild' scenarios with high-energy noise (such as strenuous motion and sudden illumination changes), as shown in Table~\ref{tab:cross3}, we used the UBFC-rPPG and PURE datasets as the combined training set and conducted tests on the VIPL-HR dataset. In this setting, Interv-rPPG improves over PhysMambaFormer* by 3.32 MAE points, which is larger than the corresponding improvement on clean-domain transfer. This gap is consistent with the design of PCP: because the editor intervenes only on the low-frequency chrominance component, training favors hypotheses that remain valid under controllable chrominance editing rather than periodic patterns that are merely dominant in a specific source dataset.
	
	\begin{figure}[t]
		\centering
		\includegraphics[scale=0.21]{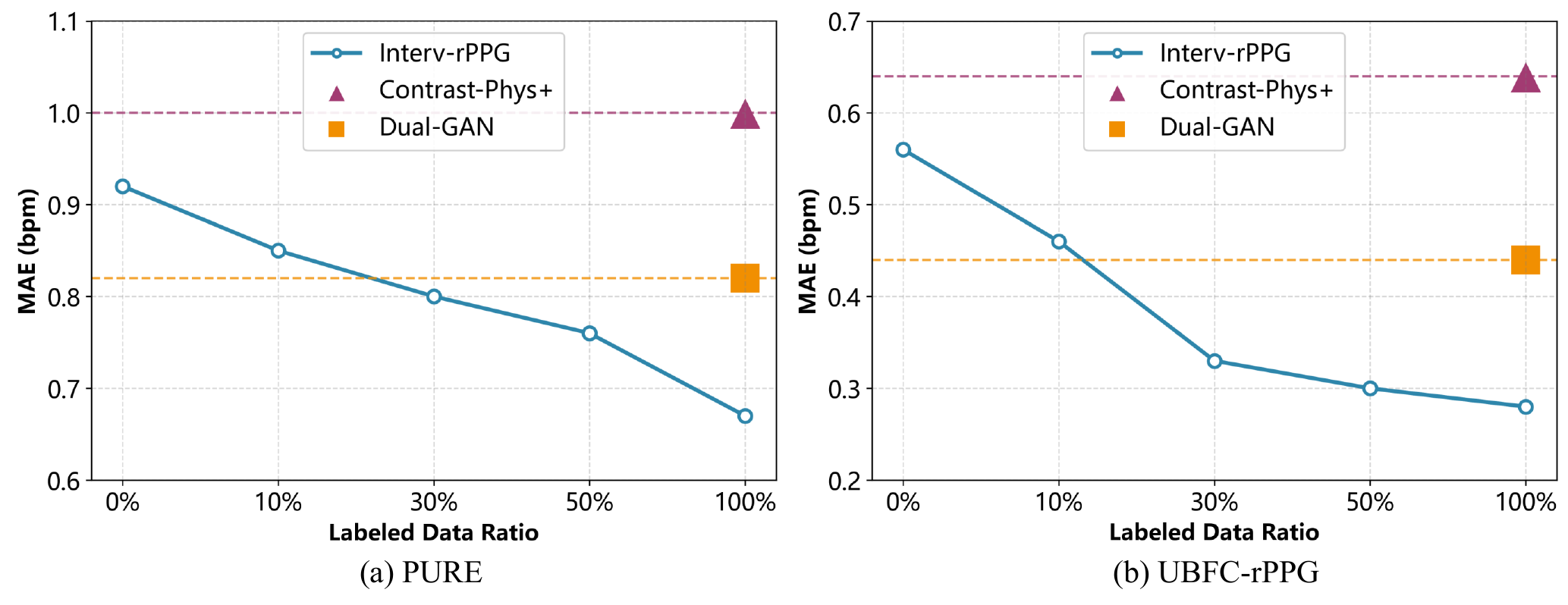}
		\vspace{-0.3em}
		\caption{\small{
				Fine-tuning results of MAE under different labeled data ratios. Fig.~\ref{fig:Finetune}(a) shows the fine-tuning results on the PURE dataset, while Fig.~\ref{fig:Finetune}(b) shows the fine-tuning results on the UBFC rPPG dataset.
			}
		}
		\label{fig:Finetune}
		\vspace{-0.3em}
	\end{figure}

	\noindent\textbf{Diagnostic analysis of nuisance sensitivity.}\quad 
	To directly test whether correlation-based SSL baselines are more easily dominated by spurious motion or illumination periodicities, we additionally applied synthetic periodic illumination flicker and rhythmic motion perturbations to held-out PURE clips and measured the change in HR estimation error. The results are summarized in Table~\ref{tab:diagnostic}.
	
	\begin{table*}[t]
		\centering
		\caption{Diagnostic nuisance-sensitivity analysis on held-out PURE clips.}
		\label{tab:diagnostic}
		\begin{threeparttable}
			\begin{tabular}{lcccc}
				\hline
				Method & Clean & +Illum. & +Motion & Avg. $\Delta$MAE \\ \hline
				SiNC \cite{speth2023non} & \textbf{0.61} & 4.11 & 4.26 & 3.57 \\
				Contrast-Phys+ \cite{sun2024contrast} & 1.00 & 3.42 & 3.95 & 2.69 \\
				PhysMambaFormer* & 0.85 & 2.36 & 2.88 & 1.77 \\
				\textbf{Interv-rPPG(Ours)} & 0.92 & \textbf{1.45} & \textbf{1.81} & \textbf{0.71} \\ \hline
			\end{tabular}
			\begin{tablenotes}
				\footnotesize
				\item Avg. $\Delta$MAE denotes the mean performance increase under the two nuisance perturbations relative to the clean setting. Lower is better.
			\end{tablenotes}
		\end{threeparttable}
	\end{table*}
	
	Conventional SSL baselines \cite{speth2023non, sun2024contrast} exhibit substantially larger error increases once nuisance periodicity is injected, indicating that their recovered signals are more coupled to dominant periodic artifacts. In contrast, Interv-rPPG shows the smallest average degradation because a hypothetical signal must remain valid under explicit chrominance intervention and verification. Compared with PhysMambaFormer*, it sacrifices only a small amount of clean-setting accuracy (0.92 vs 0.85 MAE) while achieving much stronger robustness after perturbation, which is consistent with the slight clean-domain gap but clear cross-domain advantage observed earlier. This diagnostic evidence directly supports the causal motivation of PCP and helps explain the improved transfer performance observed in Tables~\ref{tab:cross1}--\ref{tab:cross3}. The robustness difference between the SiNC and Interv-rPPG  under periodic illumination perturbation can be further intuitively verified by visualization (see Fig.~\ref{fig:illumination_visualization}).

	\begin{table*}[t]
		\centering
		\caption{Label fine-tuning experimental results under different label ratios.} 
		\label{tab:finetune}
		\resizebox{\textwidth}{!}{
		\begin{threeparttable} 
			\begin{tabular}{cccccccccc}
				\hline
				\multirow{2}{*}{Method} & \multicolumn{3}{c}{PURE}          & \multicolumn{3}{c}{UBFC-rPPG} & \multicolumn{3}{c}{VIPL-HR} \\ \cline{2-10} 
				& \thead{MAE\\(bpm)} & \thead{RMSE\\(bpm)} & r         & \thead{MAE\\(bpm)}  & \thead{RMSE\\(bpm)} & r      & \thead{MAE\\(bpm)}  & \thead{RMSE\\(bpm)} & r     \\ \hline
				Interv-rPPG(0\%)        & 0.92       & 1.52     & 0.99      & 0.56      & 1.01     & 0.99   & 4.47     & 7.26     & 0.83  \\
				Interv-rPPG(10\%)       & 0.85       & 1.46     &  0.99     &  0.46     & 0.85     & 0.99 & 4.15     &  6.98    & 0.84   \\
				Interv-rPPG(30\%)       &  0.80     &  1.39     &  0.99     &  0.33     & 0.73     & 0.99   &  4.04     &  6.82    &  0.84  \\
				Interv-rPPG(50\%)       &  0.76     &  1.22     &  0.99     & 0.30  &   0.52   &   0.99  & 4.01       &  6.63     &  0.84  \\
				Interv-rPPG(100\%)      &  0.67     &  1.04     &  0.99     & 0.28  &   0.47    &  0.99  & 3.96   &  6.02     &  0.85  \\ \hline
			\end{tabular}
		\end{threeparttable} 
	}
	\end{table*}

	\noindent\textbf{Label fine-tuning experiment.} \quad  
	To further explore the representational capability and data efficiency of our self-supervised pre-trained model, we conducted supervised fine-tuning experiments. We used the Interv-rPPG model pre-trained on all unlabeled data as the starting point, and then fine-tuned it on the training sets of UBFC-rPPG, PURE, and VIPL-HR using 10\%, 30\%, and 50\% of the labeled data, respectively. As shown in Table~\ref{tab:finetune}, we compared the fine-tuned models with the zero-shot (0\% labeled data) performance of Interv-rPPG, as well as the fully supervised Interv-rPPG trained on 100\% labeled data.
	
	Our zero-shot self-supervised model (0\% labeled data) has already achieved a performance of 0.56 MAE on UBFC-rPPG. When fine-tuned with only 10\% labeled data, the model's performance improved to 0.46 MAE. This indicates that the PCP paradigm has provided the model with a robust rPPG feature space, enabling rapid alignment to domain-specific supervised signals with minimal labeled data. With the labeled data increasing to 30\% and 50\%, the model's performance improved steadily. To compare the performance of the Interv-rPPG(0\%) trained with different ratios of labeled data against the supervised baseline model trained on full labeled data, we present Fig.~\ref{fig:Finetune}. Fig.~\ref{fig:Finetune}(a) illustrates the fine-tuning results of Interv-rPPG on the PURE dataset: when fine-tuned with only 30\% of the labeled data, the model's performance has exceeded that of Dual-GAN, which is supervisedly trained on 100\% labeled data. Fig.~\ref{fig:Finetune}(b) shows its fine-tuning performance on the UBFC-rPPG dataset: after fine-tuning with 30\% labeled data, Interv-rPPG achieves an MAE of 0.33, which is 0.11 lower than that of Dual-GAN. This result demonstrates the great value of our self-supervised pre-training: the causal representations learned through the PCP paradigm allow the model to achieve or even surpass the performance of fully supervised methods requiring only half or less of the labeled data. This is of great significance for addressing the fundamental challenge of scarce labeled data in the rPPG field.

	\begin{table}[t]
		\centering
		\caption{Ablation study for HR estimation on PURE dataset.} 
		\label{tab:ablation}
		\resizebox{0.45\textwidth}{!}{
		\begin{threeparttable} 
			\begin{tabular}{cccccc}
				\hline
				\thead{Falsifiability\\via Nulling} & \thead{Axiomatic\\Equivariance} & \thead{Spatio-temporal\\Prior}  & \thead{MAE\\(bpm)} & \thead{RMSE\\(bpm)} & r \\ \hline
				& \Checkmark                      & \Checkmark                     & 2.53   &  3.15    & 0.97   \\
				\Checkmark                        &            & \Checkmark        & 1.79   &  2.32    & 0.98   \\
				\Checkmark                        & \Checkmark          &          & 1.22   &  1.48    & 0.98   \\
				\Checkmark                 & \Checkmark       & \Checkmark         & 0.92   &  1.52    & 0.99 \\ \hline
			\end{tabular}
		\end{threeparttable} 
	}
	\end{table}

	\subsection{Ablation Study}
	\label{sec:ablation}
	To verify the necessity of each key component in the Interv-rPPG framework, we designed ablation experiments. Taking the intra-dataset HR estimation on the PURE dataset, we removed three core modules sequentially: the falsifiability via nulling module, the axiomatic equivariance module, and the spatio-temporal prior module.
	
	As shown in Table~\ref{tab:ablation}, the full Interv-rPPG model achieved the optimal performance with a 0.92 MAE. After removing the falsifiability via nulling module, the model performance degraded drastically (MAE soared to 2.53). This is the most critical module in our framework. Without it, the model loses the ability to falsify incorrect hypotheses (e.g., motion noise), the PCP closed-loop is broken, and the model falls back into the `correlation trap'. Removing the axiomatic equivariance module also led to a significant performance drop (MAE increased to 1.79). This indicates that enforcing the model to adhere to physical axioms such as amplitude, phase, and frequency is crucial for the model to learn generalizable rPPG representations that are independent of specific temporal patterns. When the spatio-temporal prior module was removed, the performance decreased moderately (MAE rose to 1.22). This demonstrates that while spatio-temporal priors (e.g., correlation with chrominance signals, facial region consistency) serve as effective constraints, their importance is secondary to the first two physically causal constraints. The ablation experiments support that falsifiability via nulling and axiomatic equivariance module are the theoretical cornerstones of our PCP paradigm and Interv-rPPG framework, and neither can be omitted.
	
	To verify the rationality of the design of PhysMambaFormer, we carried out supervised ablation experiments on the MMPD dataset. Removing temporal normalization or temporal difference raises MAE and lowers SNR, proving their role in suppressing slow interference and improving low‑SNR performance. Replacing the GTSS blocks with pure local temporal convolution causes the largest performance drop, indicating that long‑range temporal aggregation is critical for recovering weak rPPG signals from long videos. Removing the GAL module also leads to a slight performance drop. These modules work together to help the model better capture physiological features.

	\subsection{Video Physiological Signal Editing}
	\label{sec:rppgEdit}
	
	This section provides a detailed evaluation of the accuracy and video fidelity of our proposed controllable physiological signal editor in rPPG signal editing task.

	\begin{table}[t]
		\centering
		\caption{Architectural ablation of PhysMambaFormer on MMPD datasets.}
		\label{tab:extractor_arch}
		\resizebox{0.45\textwidth}{!}{
			\begin{threeparttable}
				\begin{tabular}{lcc}
					\hline
					Extractor Variant & \thead{MAE\\(bpm)} & SNR \\ 
					\hline
					w/o Temporal Normalization & 7.36 & -3.58 \\
					w/o Temporal Difference Prior & 7.14 & -3.35 \\
					Replace GTSS with temporal Conv blocks & 10.78 & -7.36 \\
					w/o GAL & 7.02 & -3.21 \\
					PhysMambaFormer* & 6.88 & 1.09 \\
					\hline
					
				\end{tabular}
			\end{threeparttable}
		}
	\end{table}

\begin{figure*}[h]
		\centering
		\includegraphics[scale=0.5]{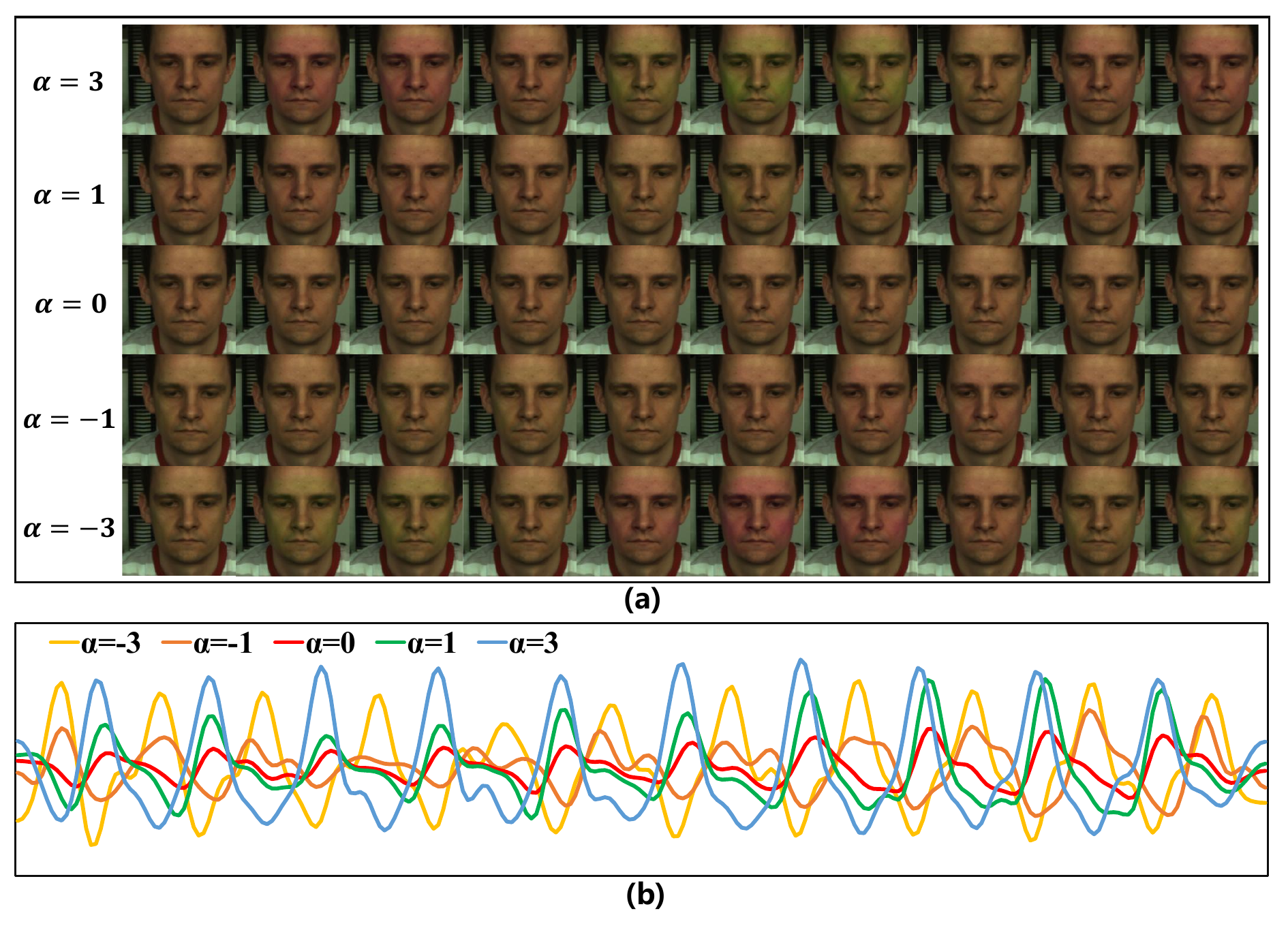}
		\vspace{-0.6em}
		\caption{\small{
				Visualization of signal amplitude editing. Fig.~\ref{fig:EditVisual}(a) illustrates the edited video frame sequences, containing 5 rows with each row corresponding to a specific $\alpha$ value. Fig.~\ref{fig:EditVisual}(b) shows the rPPG signals extracted from each video under different $\alpha$ values.
			}
		}
		\label{fig:EditVisual}
		\vspace{-0.3em}
	\end{figure*}

	\noindent\textbf{Signal amplitude editing.} \quad  
	We verified the controllable physiological signal editor's ability to precisely control the amplitude of rPPG signals. As introduced in Section~\ref{sec:Interv-rPPG}, the editor injects chrominance perturbations $\Delta C_t^K$ into the chrominance space of original video frames. To this end, we introduced a controllable parameter $\alpha$ to adjust the scaling factor of chrominance perturbations. Five values of $\alpha$ were set: $\alpha = \{3, 1, 0, -1, -3\}$, corresponding to 3× (positive superposition), 1× (positive superposition), the original video (zero perturbation), -1× (inverse superposition), and -3× (inverse superposition), respectively. These commands were applied to the same video clip, and rPPG signals were extracted from these edited videos. As shown in Fig.~\ref{fig:EditVisual}, to intuitively demonstrate the chrominance variation of videos with pulse under different $\alpha$ values, we selected a 1-second time window from the same video clip and extracted 10 visualization frames using a `select 1 frame every 2 frames' strategy, which basically covers the main phases of one cardiac cycle.

	\begin{figure*}[t]
		\centering
		\includegraphics[scale=0.31]{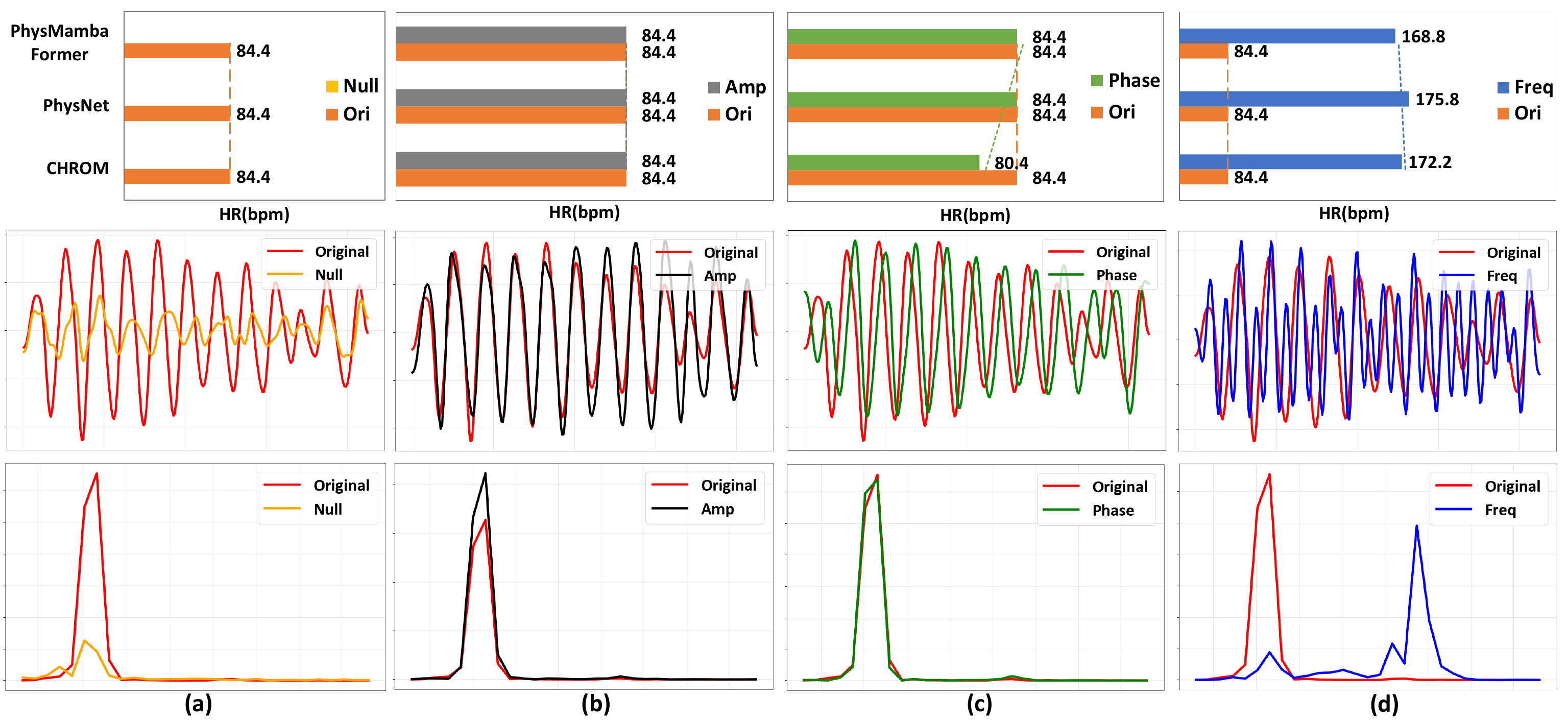}
		\vspace{-0.5em}
		\caption{\small{
				The result of the controlled physiological signal editor performing (a) HR removal, (b) amplitude intervention, (c) phase intervention, and (d) frequency intervention on the PURE dataset. The first row shows bar charts of HR values extracted from the original and edited videos using different methods (CHROM, PhysNet and PhysMambaFormer); the second row displays rPPG signal waveforms; the third row presents the power spectral density of HR.
			}
		}
		\label{fig:VisEditor}
		\vspace{-0.3em}
	\end{figure*}

	Due to the blue channel in images being highly susceptible to scattering and noise, and the red channel being affected by the mixture of epidermal and deep-layer tissues, studies \cite{ontiveros2023evaluating, gs2024remote} have indicated that the pulsation in the green channel is the clearest. Consequently, most current rPPG signal amplification algorithms \cite{takeda2019video} primarily amplify signals along the chrominance direction dominated by the green channel.

	Oxygenated and deoxygenated hemoglobin exhibit the strongest absorption in the green light band \cite{chan2023estimating}. Thus, during each systole, the increase in superficial blood volume leads to enhanced absorption of green light, resulting in a slight decrease in the intensity of green reflected light captured by the camera, which is perceived by the human eye as a redder or darker skin tone. In contrast, during diastole, the chrominance shifts back to a greener visual effect. Since the editor restricts signal intervention to the low-frequency chrominance direction orthogonal to luminance, this color principal axis is further emphasized rather than diluted by luminance or texture. When $\alpha > 0$, the facial colors of the 10 sample frames oscillate reciprocally between `light red $\rightarrow$ dark $\rightarrow$ light $\rightarrow$ dark green $\rightarrow$ light red'; when $\alpha < 0$, the color evolution direction is reversed.

	Fig.~\ref{fig:EditVisual}(b) overlays the rPPG curves under different $\alpha$ values. For $\alpha > 0$, the phase of the edited signal is consistent with the original signal, and the signal amplitude increases with the scaling factor. For $\alpha < 0$, the injected inverse chrominance perturbations induce a 180-degree phase inversion of the rPPG signal. This result demonstrates that our editor, via FiLM layers, achieves precise and linear control over both the amplitude and phase of rPPG signals.

\begin{table*}[t]
	\centering
	\caption{Quantitative controllability evaluation on UBFC-rPPG and PURE. CHROM, POS, PhysNet, and PhysFormer are used to extract HR from the edited videos. We report the MAE (bpm) between the extracted HR of the edited video and the target HR. Lower is better.}
	\label{tab:control_quant}
	\resizebox{\textwidth}{!}{
		\begin{threeparttable}
			\begin{tabular}{ccccccccc}
				\hline
				\multirow{2}{*}{Method} & \multicolumn{4}{c}{UBFC-rPPG} & \multicolumn{4}{c}{PURE} \\ \cline{2-9}
				& \thead{HR\\Removal} & \thead{Amplitude\\Intervention} & \thead{Phase\\Intervention} & \thead{Frequency\\Intervention} & \thead{HR\\Removal} & \thead{Amplitude\\Intervention} & \thead{Phase\\Intervention} & \thead{Frequency\\Intervention} \\ \hline
				CHROM \cite{de2013robust} & 5.84 & 0.62 & 0.91 & 2.46 & 4.95 & 0.58 & 0.83 & 2.31 \\
				POS \cite{wang2016algorithmic} & 6.27 & 0.74 & 0.97 & 2.71 & 5.43 & 0.66 & 0.95 & 2.54 \\
				PhysNet \cite{yu2019remote1} & 3.18 & 0.41 & 0.56 & 1.84 & 2.76 & 0.37 & 0.52 & 1.73 \\
				PhysFormer \cite{yu2022physformer} & 2.63 & 0.33 & 0.48 & 1.61 & 2.21 & 0.29 & 0.44 & 1.48 \\ \hline
			\end{tabular}
		\end{threeparttable}
	}
\end{table*}

	\begin{figure}[t]
		\centering
		\includegraphics[scale=0.32]{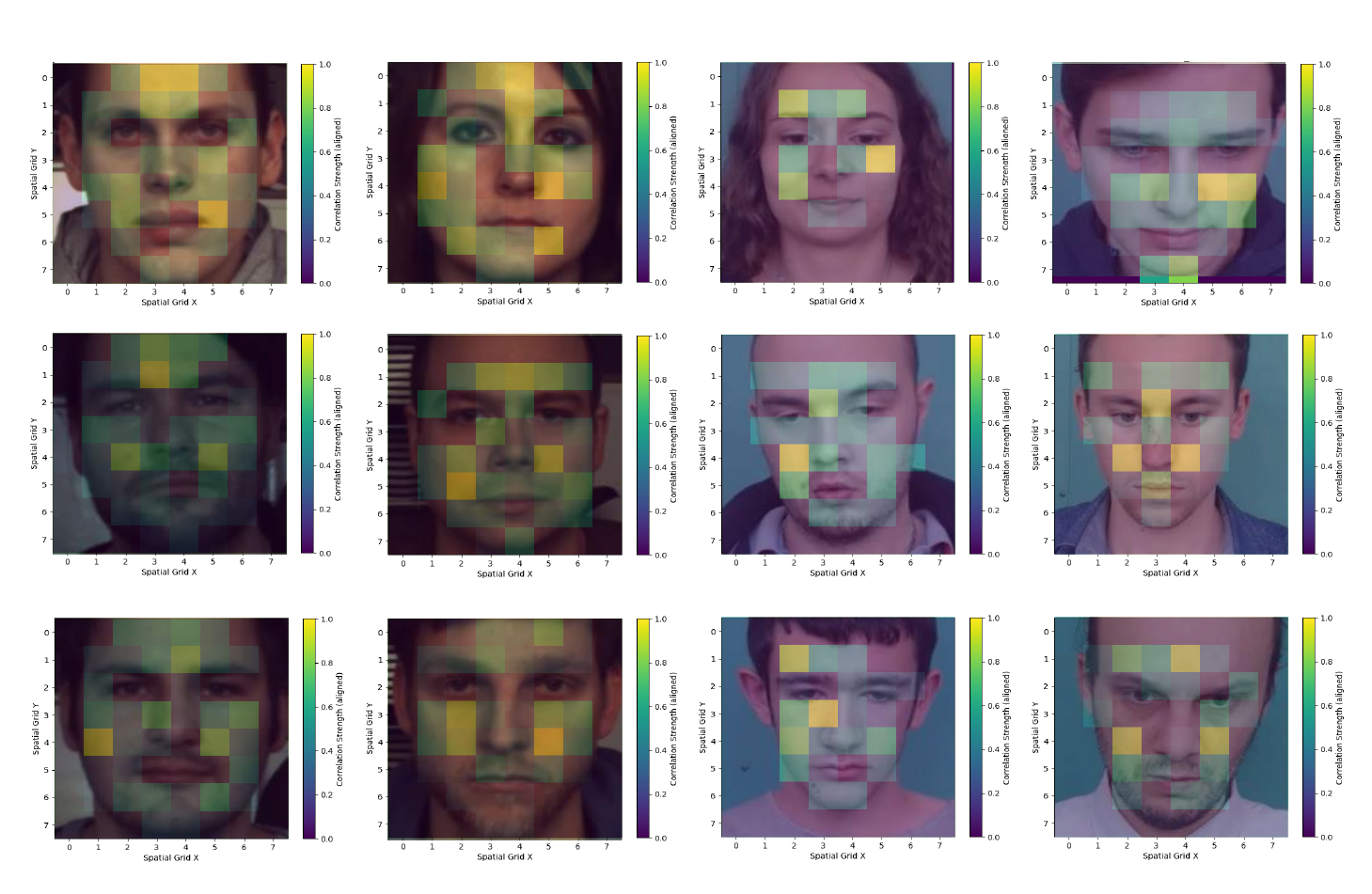}
		\vspace{-0.3em}
		\caption{\small{
				Visualization of the editing attention region ($PSM$). The bright areas (yellow) indicate where the editor concentrates its chrominance intervention ($\Delta C_t^K$), while dark areas (blue) receive near-zero attention.
			}
		}
		\label{fig:W8}
		\vspace{-0.3em}
	\end{figure}

	\vspace{0.3em}
	\noindent\textbf{Signal physical attribute editing.}\quad 
	To further verify the precise controllability of the physiological signal editor, we designed experiments for more complex physical attribute editing, including: (1) HR removal, (2) amplitude modification, (3) phase modification, and (4) frequency modification. We evaluate edited videos on the UBFC-rPPG and PURE datasets using four independent HR extractors: POS, CHROM, PhysNet, PhysFormer. For each sample, we define the target post-editing HR based on the intended operation: 0 bpm for HR removal, the original HR for amplitude and phase interventions, and $\rho \times$ original HR for frequency intervention. We then compute the MAE between the HR extracted from the edited video and the corresponding target HR. Lower MAE indicates more accurate and precise execution of the intended rPPG modification.
	
	The quantitative results is shown in Table~\ref{tab:control_quant}. Across both UBFC-rPPG and PURE, amplitude and phase intervention consistently achieve sub-1 bpm MAE for all four extractors, showing that the editor accurately preserves the cardiac frequency while only modifying the intended waveform attribute. Frequency intervention also remains highly accurate, with MAE bounded within 1.48--2.71 bpm, which confirms that the dominant spectral peak is reliably shifted to the desired target frequency. HR removal is a difficult setting because the edited video ideally contains no dominant pulse peak; nevertheless, the residual HR error is still substantially lower than the physiological HR range of the original videos, and the two deep models (PhysNet and PhysFormer) report the smallest residuals on both datasets. Thus, the consistent low errors across traditional and learning-based extractors demonstrate that the proposed editor applies the intended rPPG modifications with good accuracy and precision.

	\begin{table*}[t]
		\centering
		\caption{Comparison of video quality after editing on PURE and UBFC rPPG datasets. For Interv-rPPG, PSNR, SSIM, and LPIPS are the final averages computed over the four edited-video scenarios: HR removal, amplitude intervention, phase intervention, and frequency intervention.} 
		\label{tab:Quality_video}
		\resizebox{\textwidth}{!}{
			\begin{threeparttable}
			\begin{tabular}{ccccccc}
				\hline
				\multirow{2}{*}{Method}     & \multicolumn{3}{c}{UBFC-rPPG} & \multicolumn{3}{c}{PURE}   \\ \cline{2-7} 
				& PSNR   & SSIM  & LPIPS(1e-3)  & PSNR  & SSIM & LPIPS(1e-3) \\ \hline
				PulseEdit \cite{chen2022pulseedit}        & 52.83  & \textbf{0.99}  & -            & 57.81 & 0.97 & -           \\
				Privacy-Phys  \cite{sun2022privacy}   & -      & -     & -            & 58.38 & -    & -           \\
				PhysioLatent \cite{zhou2025editing}   & 40.04  & 0.98  & -            & 38.94 & 0.98 & -           \\
				Phys-EdiGAN  \cite{tu2025phys}   & 35.94  & 0.96  & 26.00        & 34.41 & 0.95 & 32.00       \\
				\textbf{Interv-rPPG(Ours)}   & \textbf{70.51}  & \textbf{0.99}  & \textbf{0.08}   & \textbf{72.62} & \textbf{0.99} & \textbf{0.06}        \\ \hline
			\end{tabular}
		\end{threeparttable}
		}
	\end{table*}

	Fig.~\ref{fig:VisEditor} shows the signal editing visualization of a sample from the PURE dataset. Fig.~\ref{fig:VisEditor}(a) presents the HR removal result: the edited video has much lower rPPG signal energy and HR power spectral density (PSD) than the original, verifying effective suppression of the original HR. Figs.~\ref{fig:VisEditor}(b) and (c) show the amplitude and phase modification results, respectively. As amplitude and phase changes do not alter the intrinsic HR frequency, HR values remain consistent before and after editing. The waveform in (b) has a larger amplitude, while that in (c) is phase-shifted right by half a cardiac cycle. Fig.~\ref{fig:VisEditor}(d) illustrates the result of doubling the original rPPG frequency: the edited waveform oscillates twice within one original cardiac cycle, with HR synchronously doubled. This experiment confirms that our editor can precisely control rPPG amplitude and independently intervene in its temporal attributes (phase, frequency), achieving the core design goal of axiomatic equivariance.

	\vspace{0.3em}
	\noindent\textbf{Visualization of the editing attention region.}\quad 
	We emphasize that our editing strategy relies on targeted intervention based on the Perturbation Support Map (PSM), rather than performing uniform perturbation across the entire facial region. To verify this targeting characteristic, we visualized the $PSM$ generated during the editing process.

	As illustrated in Fig.~\ref{fig:W8}, the $PSM$ attention map clearly demonstrates that our editor concentrates its core intervention (i.e., chrominance perturbation $\Delta C_t^K$) on regions with the strongest rPPG signals, such as the forehead and cheeks. These regions exhibit richer blood perfusion and are relatively less affected by motion artifacts and facial expressions. In contrast, the attention weights for regions with weak or non-existent rPPG signals (e.g., hair, eyes, and background) approach zero.

	This visualization result confirms that our editing method is spatially aware and physiologically driven, rather than implementing blind global modifications. This targeted intervention mode not only ensures the efficiency of editing operations and the fidelity of video content but also minimizes interference to irrelevant regions. This stands in stark contrast to methods like PulseEdit, which perform global perturbation on the entire Region of Interest (ROI), a design that is highly prone to introducing non-physiological artifacts and compromising the authenticity of editing results.

	\vspace{0.3em}
	\noindent\textbf{The quality of the edited video.}\quad 
	In the physiological signal editing task, maintaining video quality is one of the key metrics to evaluate the superiority of editing methods. On the premise of effectively modifying the rPPG signal of the video, higher video fidelity is preferred. We quantified the fidelity of edited videos using objective metrics, selecting Peak Signal-to-Noise Ratio (PSNR), Structural Similarity Index Measure (SSIM), and Learned Perceptual Image Patch Similarity (LPIPS) to compare the differences between edited and original videos—where a lower LPIPS value indicates higher perceptual similarity, and higher PSNR and SSIM values denote better image structure and pixel similarity.

	Four typical editing operations were selected as test cases. As shown in Table~\ref{tab:Quality_video}, the test results on the PURE and UBFC-rPPG datasets demonstrate that the Interv-rPPG signal editor outperforms other comparative editing methods across all three metrics. Specifically, on the UBFC-rPPG dataset, the PSNR of the proposed method is 17.68 higher than that of PulseEdit \cite{chen2022pulseedit}. On the PURE dataset, its PSNR and SSIM are improved by 14.81 and 0.02 compared with PulseEdit. This advantage is attributed to our Laplacian pyramid decomposition strategy: the strategy perfectly preserves all high-frequency layers of the video (including texture, edge, and identity information) and the luminance component of the low-frequency layer, only applying slight residual perturbations to the low-frequency chrominance component, thereby minimizing damage to the original video information. Furthermore, the LPIPS value of Interv-rPPG is two orders of magnitude lower than that of Phys-EdiGAN \cite{tu2025phys} on both UBFC-rPPG and PURE. As LPIPS is a core metric for measuring perceptual similarity, this result indicates that our targeted editing limited to the low-frequency chrominance domain is almost imperceptible to human visual perception.
	
	In summary, our physiological signal editor not only achieves precise and controllable physical intervention on rPPG signals but also maximally retains all non-physiology-related information in the video, attaining excellent video fidelity.

	\vspace{-0.5em}
	\section{Conclusions and Future Work}
	\vspace{-0.1em}
	\label{sec:conc}

	This paper introduces Physiological Causal Probing (PCP), a novel self-supervised rPPG paradigm that shifts learning from passive statistical correlation to active, physics-based causal intervention. By enforcing the core principles of `Falsifiability via Nulling' and `Axiomatic Equivariance', the model is compelled to differentiate true, chrominance-based physiological signals from non-cancellable artifacts. This is implemented through the Interv-rPPG framework, which pairs a PhysMambaFormer extractor for hypothesis generation with a Laplacian pyramid-based editor that executes precise, disentangled interventions on the signal's amplitude, phase, and frequency while preserving video fidelity. Experimental results validate this approach, demonstrating superior cross-domain generalization from clean to `in-the-wild' datasets. We further show that the Stage-III randomly initialized PCP extractor consistently surpasses the Stage-I supervised extractor $E^{*}$ in challenging cross-dataset settings, while a fully unlabeled Stage-II editor-training variant remains competitive.

	Looking ahead, we aim to refine the intervention mechanism. While the editor maintains high video fidelity, the intervention process itself can still introduce subtle noise, particularly on relatively clean datasets where the original interference level is limited. The remaining clean-domain gap suggests that future work should combine intervention-based invariance with even lower-disturbance representations on high-SNR data. Beyond this, future work will extend the PCP paradigm to large-scale, multi-domain, unlabeled rPPG datasets with diverse devices, motions, lighting, and skin tones, to reduce the dataset-size limitation of current public benchmarks. We are also interested in combining this direction with Test-Time Adaptation (TTA), allowing the model to continuously self-correct under unseen environmental noise during real-world deployment.

	\bibliographystyle{apalike}
	\bibliography{sn-bibliography}
	
\end{document}